\pdfoutput=1
\documentclass[11pt]{article}

\usepackage[final]{acl}

\usepackage{times}
\usepackage{latexsym}

\usepackage[T1]{fontenc}
\usepackage[utf8]{inputenc}

\usepackage{microtype}

\usepackage{inconsolata}
\usepackage{makecell}
\usepackage{times}
\usepackage{latexsym}
\usepackage{graphicx}
\usepackage{amsthm,amsmath,amssymb,bm}
\usepackage{mathrsfs}
\usepackage{booktabs}
\usepackage{algorithm}
\usepackage{algpseudocode}
\usepackage{bbm}
\usepackage{comment}
\usepackage{multirow}
\usepackage{bbding}
\usepackage{array}
\usepackage{tcolorbox}
\usepackage{colortbl}
\usepackage{xcolor}
\usepackage{xspace}
\usepackage{enumitem}
\usepackage{threeparttable}
\usepackage{CJKutf8}
\usepackage{pifont}
\newcommand{\model}[0]{\textsc{IF-Critic}\xspace}

\newcommand{\blanksymbolfootnote}[1]{%
  \renewcommand{\thefootnote}{}% Remove footnote number
  \footnote{#1}%
  \setcounter{footnote}{0} % Reset footnote counter
  \renewcommand{\thefootnote}{\arabic{footnote}}% Restore footnote number
}

\title{\model: Towards a Fine-Grained LLM Critic for Instruction-Following Evaluation}

\author{Bosi Wen$^{1,\dagger, *}$ \quad Yilin Niu$^{2,*}$\quad Cunxiang Wang$^{2}$\quad Pei Ke$^{3}$\quad Xiaoying Ling$^{2}$ \\
\textbf{Ying Zhang$^{2}$\quad  Aohan Zeng$^{4}$\quad Hongning Wang$^{1}$ \quad Minlie Huang$^{1,\ddagger}$} \\
$^1$The Conversational Artificial Intelligence (CoAI) Group, Tsinghua University \\
$^2$Zhipu AI \quad $^3$University of Electronic Science and Technology of China   \\
$^4$The Knowledge Engineering Group (KEG), Tsinghua University \\
\tt\small wbs23@mails.tsinghua.edu.cn, aihuang@tsinghua.edu.cn \\}

\begin{document}

\maketitle

\blanksymbolfootnote{$^\dagger$Work done when this author interned at Zhipu AI.}
\blanksymbolfootnote{$^*$Equal contribution}
\blanksymbolfootnote{$^\ddagger$Corresponding author}

\begin{abstract}
Instruction-following is a fundamental ability of Large Language Models (LLMs), requiring their generated outputs to follow multiple constraints imposed in input instructions. 
Numerous studies have attempted to enhance this ability through preference optimization or reinforcement learning based on reward signals from LLM-as-a-Judge. 
However, existing evaluation models for instruction-following still possess many deficiencies, such as substantial costs and unreliable assessments. 
To this end, we propose \model, an LLM critic for fine-grained, efficient, and reliable instruction-following evaluation.
We first develop a checklist generator to decompose instructions and generate constraint checklists. 
With the assistance of the checklists, we collect high-quality critique training data through a multi-stage critique filtering mechanism and employ a constraint-level preference optimization method to train \model. 
Extensive experiments show that the evaluation performance of \model can beat strong LLM-as-a-Judge baselines, including o4-mini and Gemini-3-Pro. 
With the reward signals provided by \model, LLMs can achieve substantial performance gains in instruction-following optimization under lower
computational overhead compared to strong LLM critic baselines.
Our code and model are available at \url{https://github.com/thu-coai/IF-CRITIC}.
\end{abstract}

\section{Introduction}
Large language models (LLMs) have demonstrated remarkable capabilities in various NLP tasks \cite{zhao2023survey}.
Among these, instruction-following stands out as one of the most crucial requirements for LLM applications \cite{ouyang2022training}.
In the real-world use of LLMs, nearly all tasks are formulated as instruction-following, where human instructions specify task requirements and impose corresponding constraints on the model output
\cite{jiang-etal-2024-followbench}.
Accurately following instructions serves as a key factor in ensuring the helpfulness and reliability
of LLMs \cite{huang2024survey}.

However, LLMs still fall short in instruction-following, especially when dealing with complex instructions that involve numerous constraints \cite{jiang-etal-2024-followbench}.
Many research attempts have been made
to improve the instruction-following ability of LLMs.
They often leverage LLMs to synthesize complex instructions and employ LLM-as-a-Judge to evaluate compliance with each constraint, thereby deriving reward signals to perform preference optimization or reinforcement learning \cite{zhang-etal-2025-iopo, he-etal-2024-complex, ren-etal-2025-step, liu2025recast, peng-etal-2025-verif}. 

\begin{figure}[t]
\scriptsize
    \centering
    \includegraphics[width=1.0\linewidth]{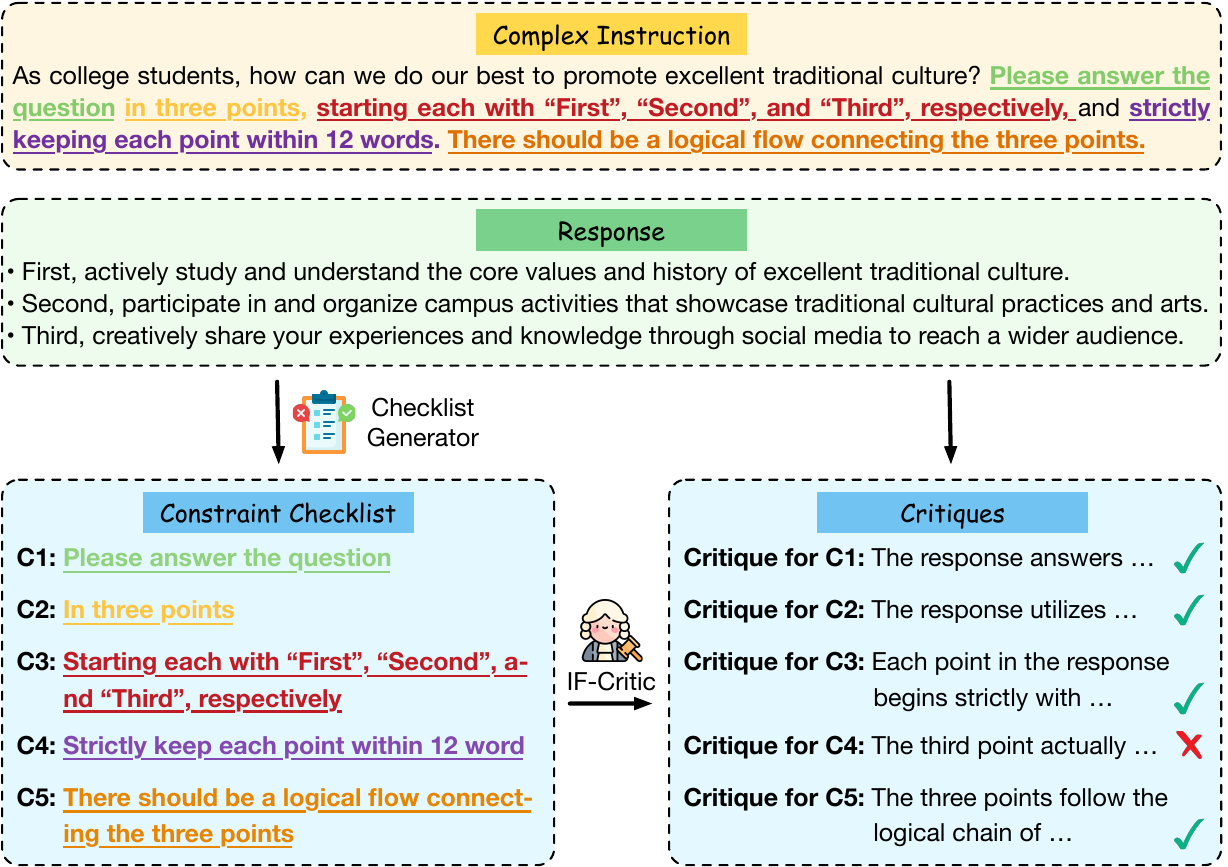}
    \caption{A usage example of \model: Given an instruction and a response, a \textit{checklist generator} first decomposes the instruction to generate a constraint checklist. Then, \model can provide fine-grained evaluations for the response with respect to its compliance with all included constraints in one inference pass.}
    \label{fig:example}
\end{figure}

Despite these efforts, the usage of LLM-as-a-Judge for instruction-following still faces two major yet underexplored challenges: 
(1) \textbf{Substantial Costs:} Existing methods typically rely on proprietary LLMs (e.g., GPT-4o) or large reasoning models (e.g., QwQ-32B) to judge each constraint in an instruction separately and use the resulting judgments as reward signals.
Since complex instructions often involve multiple constraints, these methods require repeated per-constraint inference, which is costly and difficult to scale up in model training.
(2) \textbf{Unreliable Assessments:} Current LLMs still struggle to accurately assess 
instruction-following \cite{zhang2024divide}, exhibiting low recall in error detection 
\cite{kamoi2024evaluating} and limited counting ability  \cite{zhang2024counting, fu2024large}. 
These limitations introduce noise to the evaluation and restrict the gains in model optimization.
To tackle these challenges, some works attempt to introduce handcrafted code-verifiable constraints in data construction, aiming to enhance evaluation efficiency and reliability \cite{dong2025selfplay, kim-etal-2025-systematic, li-etal-2025-ruler, peng-etal-2025-verif}. 
However, these constraints are often homogeneous and atomized, failing to cover the diversity and complexity of human instructions, such as constraint composition \cite{wen2024benchmarking}.

In this work, we propose \model, an LLM critic tailored for instruction-following evaluation. 
To enhance inference efficiency and obtain a more holistic and granular perception of the instructions, 
\model adopts a novel \textbf{checklist-guided critique generation paradigm}.
As shown in Figure \ref{fig:example}, we first utilize a \textit{checklist generator} to decompose the instructions and generate constraint checklists. 
Then, \model can evaluate compliance with all included constraints in one inference pass, eliminating the need to generate separate judgments for different constraints multiple times.
To train \model, we elaborately prompt Deepseek-R1 \cite{guo2025deepseek} to generate expert critiques guided by the checklists, then employ a \textbf{multi-stage critique filtering mechanism} to select the highest-quality critique for each constraint, 
which significantly mitigates noise and enhances the reliability of critiques, including those involving counting. 
After fine-tuning on this data, we leverage the property that the critiques contain multi-constraint evaluation results, further improving \model via a novel \textbf{constraint-level preference optimization method}.
This method performs fine-grained, constraint-level comparisons between positive and negative critiques using Direct Preference Optimization (DPO) \cite{rafailov2023direct}, reinforcing the perception of crucial preference information and leading to better alignment between \model and the expert critiques.
Empirically, \model consistently outperforms various LLM-as-a-Judge baselines on four instruction-following meta-evaluation benchmarks, including the latest state-of-the-art LLMs o4-mini \cite{jaech2024openai} and Gemini-3-Pro \cite{team2023gemini}. 

Leveraging the fine-grained evaluation results from \model as reward signals, we adapt DPO and Group Relative Policy Optimization (GRPO) \cite{shao2024deepseekmath} to improve the instruction-following ability of LLMs. 
Experiments demonstrate that \model brings substantially greater improvements over strong baselines with significantly less computational overhead, highlighting its effectiveness and practicability.
Our contributions can be summarized as follows:

\begin{itemize}
    \item We propose a novel framework for the development of instruction-following critics, including the checklist-guided critique generation paradigm, the multi-stage critique filtering mechanism for high-quality training data collection, and the constraint-level preference optimization method for model training. 
    \item We introduce \model, the first instruction-following critic developed through our framework. 
    Experiments show that \model achieves superior evaluation performance compared to strong LLM-as-a-Judge baselines, including o4-mini and Gemini-3-Pro.
    \item We validate that \model can provide scalable reward signals to improve the instruction-following abilities of LLMs using DPO and GRPO. 
    \model can bring substantially greater improvements under lower computational overhead compared to strong baselines.
\end{itemize}

\section{Related Work}
% \textbf{LLM-based Evaluation.} 
\paragraph{LLM-based Evaluation.} 
Utilizing LLMs as text evaluators has gradually become prevalent \cite{wang-etal-2023-chatgpt, chen-etal-2023-exploring-use, ke-etal-2023-decompeval} due to their advantages in flexibility, interpretability, and generalization. 
Although proprietary LLMs have demonstrated excellent evaluation capabilities \cite{zheng2023judging, wang-etal-2023-chatgpt}, their practical application is limited by issues such as high cost and potential data leakage \cite{ke-etal-2024-critiquellm}. 
To reduce the reliance on proprietary LLMs, a series of works attempt to train smaller LLMs specialized in evaluation using the data generated from humans or proprietary LLMs \cite{kim2024prometheus, li2024generative, ke-etal-2024-critiquellm, hu-etal-2024-themis, kim-etal-2024-prometheus}. 
Distinct from these models that focus on evaluating overall text quality, we develop an evaluation model that provides constraint-level, fine-grained assessments of instruction-following and leverages them as reward signals to enhance the instruction-following ability of LLMs.

\paragraph{Checklist-based Evaluation.} 
Recently, using checklists for evaluation has emerged as a promising paradigm for reward modeling and has been widely applied across diverse domains, including instruction following \cite{peng2025verif, liu2025recast, qin2025incentivizing, viswanathan2025checklists, huang2025reinforcement}, reasoning \cite{zhou2025breaking, sanders2026generating}, and long-form generation \cite{gunjal2025rubrics, shao2025dr}. 
This paradigm associates each instruction with a list of verifiable checklist items and derives the reward for a response based on the proportion of satisfied items.
Through fine-grained verification of each item to calculate the final reward, this approach alleviates the problem of overly coarse supervision \cite{liu2025openrubrics} and improves evaluation robustness \cite{shen2026rethinking}.
However, current work typically relies on proprietary LLMs or large reasoning models to separately verify each checklist item, which is costly and difficult to scale. 
To address this scaling bottleneck, we develop a fine-grained LLM critic to efficiently provide reliable verification results for checklist items.

\paragraph{Instruction-following Optimization.} 
As LLMs are increasingly applied to real-world complex tasks, the ability to follow complex instructions with multiple constraints emerges as a key determinant of their practical utility \cite{liu2023trustworthy, lou2024large}, driving extensive studies to enhance this ability of LLMs. 
They typically leverage LLMs to synthesize complex instructions and perform preference optimization or reinforcement learning based on the reward signals from LLM-as-a-Judge \cite{sun2024conifer, zhang-etal-2025-iopo, cheng2025spar, ren-etal-2025-step, liu2025recast, peng-etal-2025-verif}. 
Specifically, SPaR \cite{cheng2025spar} employs self-play tree search refinement through a fine-tuned refiner to construct DPO training data.
RECAST \cite{liu2025recast} categorizes soft and hard constraints, leveraging GPT-4o and code execution, respectively, to obtain evaluation results for GRPO training. 
However, the effectiveness of existing methods is restricted by the high costs and limited instruction-following evaluation ability of LLM-as-a-Judge, and code-verifiable constraints fail to cover the diversity and complexity of human instructions.

% \section{\model Framework}
\section{Methodology}
\begin{figure*}[!t]
\scriptsize
    \centering
    \includegraphics[width=1\textwidth]{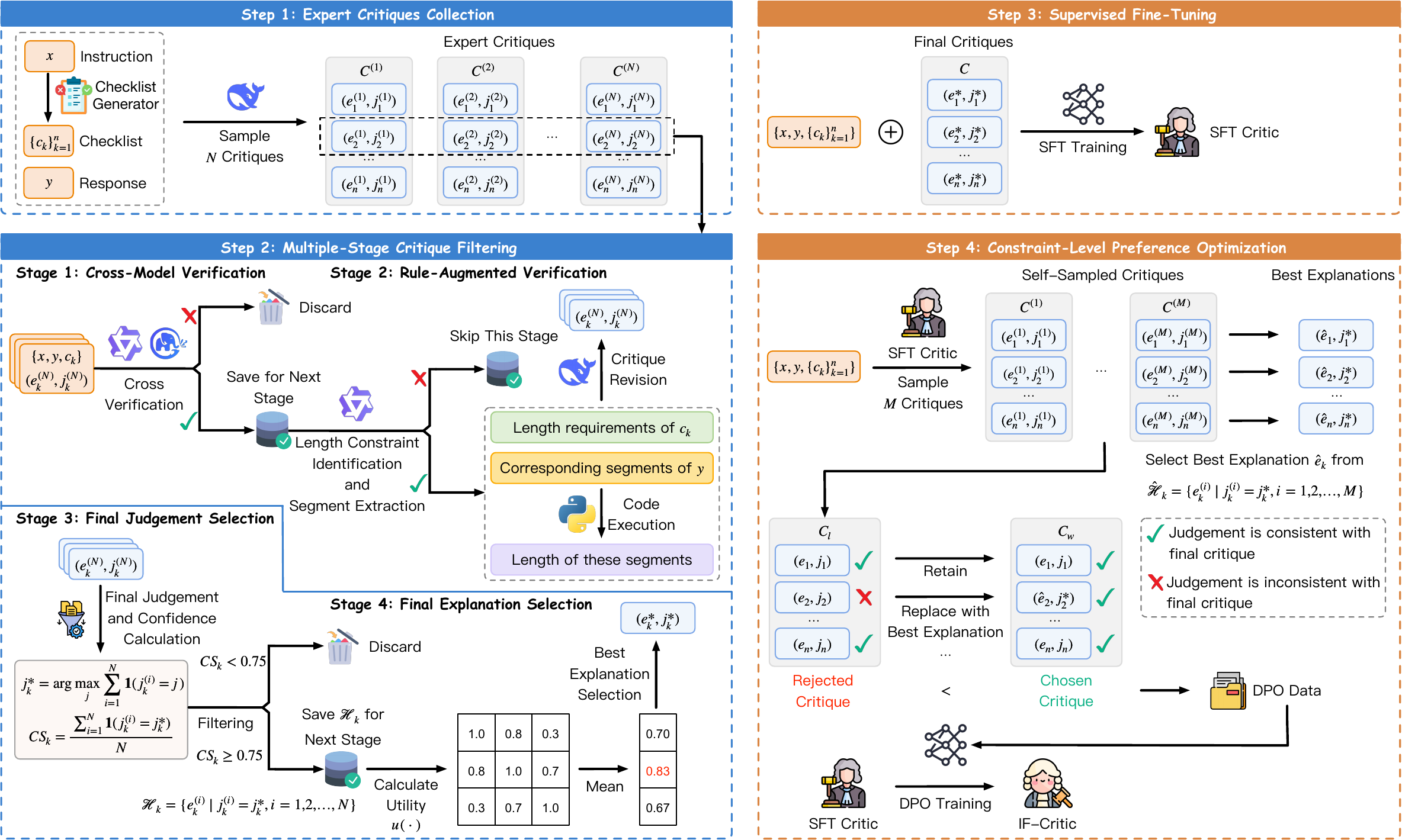}
    \caption{The pipeline of \model development. The left section illustrates the process of critique training data construction, while the right section presents the process of training \model.}
    \label{fig:overview}
\end{figure*}

\subsection{Task Definition and Method Overview}
\label{overview}
Distinct from previous LLM critics that typically generate evaluation results for a single criterion or question \cite{vu-etal-2024-foundational, wang-etal-2025-direct-judgement, anugraha2025r3}, \model leverages a constraint checklist to guide critique generation and evaluates compliance with all included constraints in one inference pass. 
Specifically, given an instruction $x$, corresponding response $y$, and a constraint checklist $\{c_k\}^n_{k=1}$ which contains all constraints within $x$, \model can provide a critique $C=\bigcup_{k=1}^{n} (e_k, j_k)$ for $y$, 
where each segment $(e_k, j_k)$ consists of an explanation $e_k$ and a binary judgment $j_k$ (0 or 1) for compliance with the respective constraint. 
This paradigm not only enhances inference efficiency but also equips \model with a more holistic and granular perception of the instructions, thus improving the reliability of instruction-following assessments.

The development of \model consists of the following steps: After collecting complex user instructions and LLM-generated responses, we first develop a checklist generator to decompose the instructions and generate constraint checklists.
With the assistance of the checklists, we elaborately prompt Deepseek-R1 to generate expert critiques and employ a multi-stage critique filtering mechanism to further enhance their quality.
Finally, we utilize the collected expert critiques and perform a constraint-level performance optimization method to train \model. 
The overall pipeline is illustrated in Figure \ref{fig:overview}.
\model can be leveraged to provide reliable reward signals, improving the instruction-following ability of LLMs via DPO or GRPO, as validated in our experiments.

\subsection{Instruction and Response Collection}
\label{instruction_and_response_collection}
To obtain diverse instructions for critic training, we collect instructions from real-world application scenarios and utilize LLMs to categorize them according to the task taxonomy of CritiqueLLM \cite{ke-etal-2024-critiquellm}, which encompasses 10 categories and covers diverse NLP applications in real-world scenarios.
To improve data quality, we prompt LLMs to score instruction quality and develop a classifier to assess their constraint complexity, which is trained with a small amount of human annotation data.
Finally, we balance the amount of instruction across task categories and acquire 55K high-quality instructions with relatively high complexity.

Then, we collect LLM responses from 15 representative models, which possess varying abilities in instruction-following.
The model list is provided in Appendix \ref{app:model_list}.
For each instruction, we randomly select two models to generate responses, resulting in a total of 110k evaluation inputs, each comprising an instruction and its corresponding LLM response.

\subsection{Critique Training Data Construction}
\label{data_costruction}

To assist \model in fine-grained evaluation, we first develop a checklist generator to decompose the constraints within instruction $x$ and generate a constraint checklist $\{c_k\}^n_{k=1}$.
To achieve this, we leverage a powerful LLM, Deepseek-R1\footnote{https://huggingface.co/deepseek-ai/DeepSeek-R1} for automatic checklist annotation in our collected instructions, and then utilize these checklists to fine-tune a base model as our checklist generator, which supports efficient checklist generation after deployment.
Manual inspection of 1,000 test samples for the generator shows that 99.29\% of generated constraints and 97.50\% of entire checklists are correct, indicating its reliability.

With the assistance of the checklists provided by our checklist generator, we proceed to employ Deepseek-R1 for expert critique annotation via a carefully designed prompt, which elicits a concise and specific explanation before the judgment for each constraint. 
Detailed prompts are provided in the Appendix \ref{app:prompt_templates}.
For each response, we collect $N$ expert critiques and apply a multi-stage critique filtering mechanism to select the highest-quality critique for each constraint in the checklist, which is then used for training \model.

The mechanism comprises four stages: cross-model verification, rule-augmented verification, final judgment selection, and final explanation selection.
The first two stages aim to filter out noise in critiques caused by two major limitations of LLM-as-a-Judge: potential evaluation bias and limited counting ability, respectively. 
The remaining two stages then select the highest-quality judgements and explanations from the filtered critiques based on consistency, further enhancing their reliability.

\paragraph{Cross-Model Verification.}
Previous works point out that LLM-as-a-Judge may suffer from potential biases \cite{zheng2023judging, ye2025justice}, undermining the reliability of its generated critiques. 
To mitigate this, we utilize two additional LLMs, GLM-4-Plus \cite{glm2024chatglm} and Qwen2.5-72B-Instruct \cite{yang2024qwen2}, to independently verify two aspects of the expert critiques for each constraint: 
(1) The correctness of the explanation, 
and (2) The consistency between the explanation and judgment. 
Detailed prompts can be found in Appendix \ref{app:prompt_templates}. 
Any critiques that fail verification by any model in any aspect are strictly filtered, accounting for 11.3\% of our data.

\paragraph{Rule-Augmented Verification.}
Considering that LLMs often struggle with accurate counting \cite{zhang2024counting, fu2024large}, we introduce rule-augmented verification to enhance the reliability of critiques involving length constraints (e.g., \textit{each title must be 8 characters long}). 
Methodologically, we first prompt Qwen2.5-72B-Instruct to identify length constraints and extract relevant segments from the responses. 
Subsequently, we employ Python scripts to identify length information within these segments automatically.
This information then guides Deepseek-R1 to revise its critiques. 
The prompts for these steps are in the Appendix \ref{app:prompt_templates}.
This approach integrates the flexibility of LLM-based evaluation with the precision of rule-based evaluation, thereby effectively handling the diverse length constraints in human instructions.

\paragraph{Final Judgement Selection.}
Given that critique generation is analogous to a Chain-of-Thought (CoT) \cite{wei2022chain} reasoning process, we incorporate self-consistency \cite{wang2023selfconsistency}, a widely used decoding strategy in CoT reasoning, to select final judgments and improve their correctness.
Specifically, for each constraint $c_k$, we select the final judgment $j_k^*$ via majority voting among multiple expert critiques. 
Furthermore, previous works show that \textit{confidence}, which is calculated as the proportion of the majority result to the total samples, is an important indicator of answer reliability \cite{xiong2024can, hu-etal-2024-themis, prasad2024self}, with higher confidence generally implying greater reliability.
Inspired by this, we discard the final critiques of constraints with confidence lower than 0.75 in our experiment, which empirically balances data retention and reliability.

\paragraph{Final Explanation Selection.}
After obtaining the correct final judgments, the last step is to select the highest-quality explanations that support these judgments. 
To achieve this, we devise a selection strategy inspired by Minimum Bayes Risk Decoding \cite{bickel2015mathematical}: For each constraint $c_k$, we first construct a hypothesis set $\mathcal{H}_k$ containing all explanations whose corresponding judgments align with the final judgments $j_k^*$.  
Then, we utilize a text similarity measure $u(\cdot)$ to calculate the consistency between each pair of explanations in $\mathcal{H}_k$. 
The explanation with the highest average consistency among others is selected as the final explanation $e_k^*$. 
The formal definition is as follows:
\begin{equation}
\small
\label{eq:hyp}
\begin{aligned}
\mathcal{H}_k = \{e_k^{(i)} \mid  j_k^{(i)}=j_k^*, i=1,2,…,N\}
\end{aligned}
\end{equation}
\begin{equation}
\small
\begin{aligned}
\label{eq:mbr}
e_k^* = \arg\max_{e \in \mathcal{H}_k} \frac{1}{|\mathcal{H}_k|} \sum_{\widetilde{e} \in \mathcal{H}_k}u(\widetilde{e}, e)
\end{aligned}
\end{equation}
Through these rigorous filtering processes, we manually inspect the quality of 70 final critiques with 353 constraints and find that 96.03\% judgments and 92.35\% explanations are completely correct, demonstrating their high quality.

\subsection{\model Training}
\label{preference_learning}
After constructing the final critique training dataset $D = \{C_i = \bigcup_{k=1}^{n_{i}} (e_{ik}^*, j_{ik}^*)\}_{i=1}^{|D|} $, 
we partition $D$ into two subsets $D_{\text{sft}} \cup D_{\text{ref}}$ and employ a two-stage paradigm to train \model, comprising supervised fine-tuning (SFT) and constraint-level preference optimization.
In the first stage, we utilize a critique generation prompt $p$ to concatenate the instruction $x$, response $y$, and checklist $\{c_k\}^n_{k=1}$, applying a standard SFT loss on vanilla LLM $\theta$:
\begin{equation}
\small
\begin{aligned}
\mathcal{L}_{\text{SFT}} = -\sum_{i=1}^{|D_{\text{sft}}|} \text{log}P_{\theta }(C_i|p_i)
\end{aligned}
\end{equation}
After obtaining the SFT critic, we further enhance its ability
via preference optimization.
Existing methods usually sample multiple independent responses from the policy model and utilize response-level rewards to select chosen and rejected examples \cite{yuan2024selfrewarding, ye2025learning, yu2025improve}.
However, since the critiques in our task involve multiple segments for different constraints, only the segments where chosen and rejected critiques differ in judgments encapsulate crucial preference information. 
Other segments 
may introduce unnecessary interference and dilute important supervision signals, leading to insufficient optimization \cite{chen-etal-2024-improving, cheng2025spar}.

To mitigate this, we introduce fine-grained, constraint-level comparisons in preference pairs construction, thus reinforcing the perception of crucial preference information.
Specifically, for each evaluation input $p$ in $D_{\text{ref}}$, we first sample $M$ critiques from the SFT critic. 
Then, we identify the best self-sampled explanation $\hat{e}_k$ for each constraint $c_k$ whose judgements align with the expert final judgement $j^{*}_k$ in a manner analogous to Equation \ref{eq:hyp} and \ref{eq:mbr} (if any).
Finally, self-sampled critiques with at least one judgment that misaligns with the expert final critique are selected as rejected critiques $C_l=\bigcup_{k=1}^{n} (e_k, j_k)$.
Retaining the segments that align with the expert final critique while replacing other misaligned segments with the best self-sampled explanations $\hat{e}_k$ and expert final judgement $j^{*}_k$ to construct chosen critiques $C_w$:
\begin{equation}
\small
\begin{aligned}
C_w =\bigcup_{k=1}^{n} \begin{cases}
  (e_k,j_k), & j_k=j^*_k \\
  (\hat{e}_k,j^*_k),  & j_k \neq j^*_k
\end{cases}
\end{aligned}
\end{equation}
Rejected critiques are discarded when some misaligned segments lack substitute best self-sampled explanations. 
Consequently, we construct a preference dataset $D_{\text{dpo}}$ where the preference pairs only differ in the segments with inconsistent judgments, thereby reinforcing crucial preference information.
Moreover, self-sampled critiques are closer to the SFT critic decoding space than the expert critique, which is generally beneficial for optimization \cite{guo2024direct}. 
Finally, we conduct preference optimization via the following DPO loss:
\begin{equation}
\small
\begin{aligned}
\mathcal{L}_{\text{DPO}}(\pi_\theta;\pi_\text{ref}) = & 
-\mathbb{E}_{(p,C_l,C_w) \sim D_{\text{dpo}}}[\log \sigma(\beta \log \frac{\pi_\theta(C_w|p)}{\pi_\text{ref}(C_w|p)} \\ & - \beta \log \frac{\pi_\theta(C_l|p)}{\pi_\text{ref}(C_l|p)})]
\end{aligned}
\end{equation}
where both $\pi_\theta$ (i.e., the trained policy model) and $\pi_\text{ref}$ (i.e., the fixed reference model) are initialized from the SFT critic.
\subsection{Evaluation Results Utilization}
\label{model_optimization}
Leveraging the evaluation results of \model as reward signals, we can apply DPO or GRPO algorithms to improve the instruction-following abilities of LLMs. 
For every input instruction $x$, the checklist generator produces the
constraint checklist $\{c_k\}^n_{k=1}$, and the policy model $\pi_\theta$ samples $K$ candidate responses $\{y_i\}^K_{i=1}$. 
Then, the reward $r_i$ for each response $y_i$ can be determined by the generated critique $C_i=\bigcup_{k=1}^{n} (e_{ik}, j_{ik})$ for $y_i$: 
\begin{equation}
\small
\label{reward}
\begin{aligned}
r_i = \frac{1}{n}\sum_{k=1}^nj_{ik}
\end{aligned}
\end{equation}
Then, we can use this reward to construct DPO training data or to conduct GRPO training.

\section{Experiments}
\label{sec:experiments}
\definecolor{c2}{RGB}{223,246,245}
\definecolor{c3}{RGB}{224,222,241}

\begin{table*} [!t]
\centering
\resizebox{\textwidth}{!} {
\small
\begin{tabular}{cl|cc|cc|cc|cc|c}
\toprule
\multirow{3}{*}{\textbf{Prompt}}  & \multirow{3}{*}{\textbf{Model}} & \multicolumn{2}{c|}{\textbf{EvalCritic}} & \multicolumn{2}{c|}{\textbf{CFBench}} & \multicolumn{2}{c|}{\textbf{TRACE}} & \multicolumn{2}{c|}{\textbf{Multi-IF}} & \textbf{Avg.} \\
\cmidrule{3-11}
 &  & Positive   & Negative   & Positive   & Negative  &  Positive   & Negative & Positive   & Negative & Average \\
 & & F1 & F1  & F1 & F1  & F1 & F1 & F1 & F1 & F1 \\
\midrule
\multirow{8}{*}{Constraint-Level} & Gemini-3-Pro & 0.887 & 0.744 &  0.934 & \underline{0.794} & \underline{0.932} & \underline{0.704} & \underline{0.962} & \underline{0.898} &  \underline{0.857} \\
& o4-mini & 0.887  & 0.717 &  0.927   & 0.755 & 0.923 & 0.599 & 0.960 & 0.890 &  0.832 \\
& GPT-4.1 & 0.883  & 0.608 &   0.919   & 0.689 &  0.924 & 0.597  & 0.932 & 0.801 &  0.794 \\
% & GLM-4-Plus & 0.873  & 0.526 &   0.900   & 0.563 &  0.926 & 0.518 & 0.913 & 0.682 & 0.738 \\
\cmidrule{2-11}
& Deepseek-R1  & 0.863   & 0.686 & 0.905 & 0.718 &  0.883 & 0.555  & 0.933 & 0.824 &  0.796 \\
& QwQ-32B & 0.856 & 0.675 & 0.900 & 0.684 & 0.901 & 0.595 & 0.931 & 0.795 &  0.792 \\
& Qwen-2.5-72B-Instruct & 0.854 & 0.362  & 0.891  & 0.459 & 0.915 & 0.374 & 0.896 & 0.590 &  0.668 \\
& Llama-3.3-70B-Instruct & 0.869 & 0.486 & 0.904  & 0.612  & 0.916 & 0.426 & 0.921 & 0.756 &  0.736 \\
& GLM-4-32B-0414 &  0.867 & 0.451  & 0.900 & 0.531  & 0.913 & 0.384  & 0.898 & 0.610 &  0.694 \\

\midrule
\midrule

\multirow{10}{*}{Checklist-Level} & Gemini-3-Pro & 0.902 & 0.742 & \textbf{0.945} & \textbf{0.808} & \underline{0.932} & 0.655 & 0.960 & 0.892 &  0.855 \\
& o4-mini & \underline{0.911}  & \underline{0.753} &  0.935  & 0.761 & 0.925 & 0.639 & \textbf{0.964} & \textbf{0.900} &  0.849 \\
& GPT-4.1 & 0.883  & 0.560 &   0.918   & 0.637 &  0.923 & 0.517 & 0.933 & 0.799 &  0.771 \\

\cmidrule{2-11}
& Deepseek-R1  & 0.895   & 0.716 & 0.928   & 0.725 &  0.910 & 0.579 & 0.938 & 0.828 &  0.815 \\
& QwQ-32B & 0.892  & 0.663  & 0.922 & 0.716  & 0.922 & 0.570 & 0.933 & 0.793 &  0.801 \\
& Qwen-2.5-72B-Instruct & 0.849 & 0.278  & 0.894  & 0.424 & 0.914 & 0.330 & 0.889 & 0.548 &  0.641 \\
& Llama-3.3-70B-Instruct & 0.862 & 0.391 & 0.905  & 0.585  & 0.915 & 0.400 & 0.916 & 0.718 &  0.712 \\
& GLM-4-32B-0414 &  0.855 & 0.347  & 0.897 & 0.512  & 0.909 & 0.325 & 0.909 & 0.664 &  0.677 \\

\cmidrule{2-11}
 & \model-14B (Ours) &  \textbf{0.927}  & \textbf{0.806} & \underline{0.937}  & 0.785 & \textbf{0.942} & \textbf{0.739}  & 0.945 & 0.844 &  \textbf{0.866} \\
\bottomrule
\end{tabular}
}
\caption{Positive F1 and Negative F1 scores of constraint-following verification on the instruction-following evaluation task. The best performance among these models is \textbf{bold}, while the second-best performance is \underline{underlined}.}
\label{tab:critic_result}
\end{table*}
\subsection{Instruction-Following Evaluation}
\paragraph{Evaluation Benchmarks.}
Given the scarcity of existing instruction-following meta-evaluation datasets, we sample instructions from four benchmarks: \textbf{EvalCritic}, \textbf{CFBench} \cite{zhang2024cfbench}, \textbf{TRACE} \cite{zhang-etal-2025-iopo}, and \textbf{Multi-IF} \cite{he2024multi} to construct meta-evaluation datasets.
EvalCritic is a random split of the dataset we used for instruction-following optimization (§\ref{instruction_following_optimization})
, whose distribution of instructions differs from \model training data.
For the first three benchmarks, we randomly sample 150 instructions.
For Multi-IF, we select the first round from the Chinese version, comprising 454 instructions.
Responses for each instruction are generated by 2 of 6 LLMs with varying capabilities.
We use our checklist generator to generate constraint checklists for EvalCritic and directly adapt the original checklists for other benchmarks.
For the first three benchmarks, two annotators independently judge compliance with each constraint, with a third inspector conducting cross-validation. 
Any discrepancies will be discussed to reach a consensus.
For Multi-IF, we use the provided verification code for evaluation.
Across these benchmarks, there are 1411 / 1070 / 1634 / 1220 constraints, respectively, of which 406 / 252 / 297 / 320 are labeled as "not followed".
\paragraph{Baselines.}
We choose state-of-the-art general and evaluation-specific LLMs as our baselines, which are listed in Appendix \ref{app:baseline_models}. 
General LLMs can serve as evaluators for instruction-following via elaborately prompting. 
We explore two prompting strategies: 
(1) \textbf{Checklist-Level Prompt}: evaluate all constraints in the given checklist within a single inference pass, which is the same as \model. 
(2) \textbf{Constraint-Level Prompt}: evaluate each constraint individually.
Evaluation-specific LLMs are mainly designed for pairwise comparison. 
To accommodate this evaluation setting, we convert the above datasets into pairwise versions. 
Specifically, for two responses generated from the same instruction, we select pairs in which one response follows all constraints while the other does not to construct pairwise samples. 
The rewards of \model\ according to Equation \ref{reward} for both responses are converted into pairwise comparisons to calculate the agreement rate. 
To ensure a fair comparison, we remove samples that result in a tie for \model.
Greedy search decoding is used for reproducibility, which has a negligible impact on the evaluation performance of reasoning models, as verified in Appendix \ref{app:decoding_strategies}.
\paragraph{Implementation Details.}
We choose Qwen-2.5-14B-Instruct \cite{yang2024qwen2} as the base model for both the checklist generator and \model.
We also experiment with other base models, and the results are provided in Appendix \ref{app:base_model}.  
Deepseek-R1 generates $N = 5$ expert critiques per response. 
During final explanation selection, we use \texttt{difflab}\footnote{https://docs.python.org/3/library/difflib.html} to implement the text similarity measure $u(\cdot)$.
During \model training, the ratio of $D_\text{sft}$ to $D_\text{ref}$ is set to 6:4. 
The number of self-sampled critiques, $M$, is set to 10, and up to 1 preference pair is constructed for each evaluation input.
More implementation details are in Appendix \ref{app:training_details}.

\begin{table} [!t]
\centering
\resizebox{\linewidth}{!} {
\begin{tabular}{l|c|c|c|c}
\toprule
\textbf{Model} & \textbf{EvalCritic} & \textbf{CFBench}  & \textbf{TRACE} & \textbf{Multi-IF} \\
\midrule
Skywork-Reward-V2-Llama-3.1-8B-40M & 0.571 & 0.689  &  0.671 & 0.705 \\
Prometheus-BGB-8x7B & 0.464 & 0.556 & 0.529 & 0.489 \\
RM-R1-Deepseek-Distilled-Qwen-32B & 0.607 & 0.711  &  0.765  & 0.625 \\
\textsc{R}RM-32B & 0.534 & 0.689  &  0.765 & 0.636 \\
\model-14B (Ours) & \textbf{0.964}& \textbf{0.956}  &  \textbf{0.882} & \textbf{0.977} \\
\bottomrule
\end{tabular}
}
\caption{Pairwise agreement rates on the instruction-following evaluation task.}
\label{tab:pairwise_result}
\end{table}
\paragraph{Main Results.}
Table \ref{tab:critic_result} presents the F1-scores for both positive and negative classes in the instruction-following evaluation task of general LLMs, while Table \ref{tab:pairwise_result} shows the pairwise agreement for evaluation-specific LLMs.
\textbf{Firstly}, \model demonstrates strong performance on various benchmarks, outperforming all baselines on average F1, including the latest state-of-the-art LLM o4-mini and Gemini-3-Pro. 
In contrast, the performance of existing general LLMs remains limited, with relatively low negative F1.  
This aligns with our motivation for developing \model to enable more reliable evaluation.
\textbf{Secondly}, existing evaluation-specific LLMs still perform poorly in evaluating instruction-following and lag far behind \model, highlighting the necessity of a critic specifically designed for this task.
\textbf{Finally}, 
reasoning models (e.g, o4-mini, DeepseekR1, and QwQ-32B) typically perform better with \textbf{Checklist-Level Prompt}, 
while other LLMs benefit more from \textbf{Constraint-Level Prompt}. 
We attribute this to two factors: 
(1) Evaluating multiple constraints at one inference pass increases task complexity. 
(2) Long-chain reasoning enables a more holistic perception of the relationships between constraints with the help of checklists. 
This observation validates the rationale of our checklist-guided critique generation paradigm.

\begin{table} [!t]
\centering
\resizebox{\linewidth}{!} {
\begin{tabular}{l|c|c|c|c}
\toprule
\textbf{Method} & \textbf{EvalCritic} & \textbf{CFBench}  & \textbf{TRACE} & \textbf{Multi-IF} \\
\midrule
\model-14B & \textbf{0.861} & \textbf{0.863}   &  \textbf{0.840} & \textbf{0.895} \\
\midrule
\multicolumn{5}{c}{\textit{Data Format}} \\
\midrule
w/ Constraint-Level Critique & 0.844  &  0.830   &  0.816 & 0.859 \\
\midrule
\multicolumn{5}{c}{\textit{Critique Filtering}} \\
\midrule
w/ Raw Data & 0.814  &  0.792  &  0.774  & 0.780 \\
w/o Cross-Model Verification & 0.851  &  0.858   &  0.832 &  0.874 \\
w/o Rule-Augmented Verification & 0.827  &  0.823   &  0.789 & 0.825 \\
w/o Final Judgement Selection & 0.840  & 0.804   &  0.821 & 0.849 \\
w/o Final Explanation Selection & 0.840  &  0.846   &  0.807  & 0.858 \\
\midrule
\multicolumn{5}{c}{\textit{Preference Learning}} \\
\midrule
w/ Vanilla DPO  & 0.797  & 0.797    &  0.785 & 0.841 \\
w/ Expert Critique   & 0.828  &  0.836  &  0.801 & 0.840 \\
w/o Preference Learning & 0.815  &  0.810  &  0.810 & 0.841 \\

\bottomrule
\end{tabular}
}
\caption{The average of Positive F1 and Negative F1 scores under different ablation settings for \model.}
\label{tab:critic_ablation}
\end{table}
\paragraph{Ablation Study.}
To investigate the impact of each design component of our framework, we conduct additional ablation studies. 
For the training data format, we decompose expert critiques to train an alternative model that generates critiques for individual constraints. 
Table \ref{tab:critic_ablation} shows that the performance of \model declines across all benchmarks, demonstrating that checklist-guided training can facilitate instruction perception and contribute to the final performance.

For the critique filtering mechanism, we remove each step individually to explore its impact on evaluation performance. 
Table \ref{tab:critic_ablation} shows that all steps contribute to the final performance, but removing rule-augmented verification yields the largest performance drop, underscoring its key role in improving discriminability for length constraints.
Notably, directly using all raw training data without critique filtering leads to the most severe performance decline, highlighting its effectiveness. 

For the preference optimization method, we explored three variants of our method: 
(1) Constructing preference pairs directly based on the agreement rates between expert critiques, denoted as Vanilla DPO,
% of multiple self-sampled critiques; 
(2) Replacing misaligned segments of rejected critiques with expert critiques rather than the best self-sampled ones,
and (3) Performing SFT on the entire training dataset without preference optimization.
Table \ref{tab:critic_ablation} shows that our constraint-level preference optimization method achieves the best performance, evidencing its effectiveness in enabling better alignment with expert critiques.

\begin{figure}[!t]
  \centering
  \includegraphics[width=1.0\linewidth]{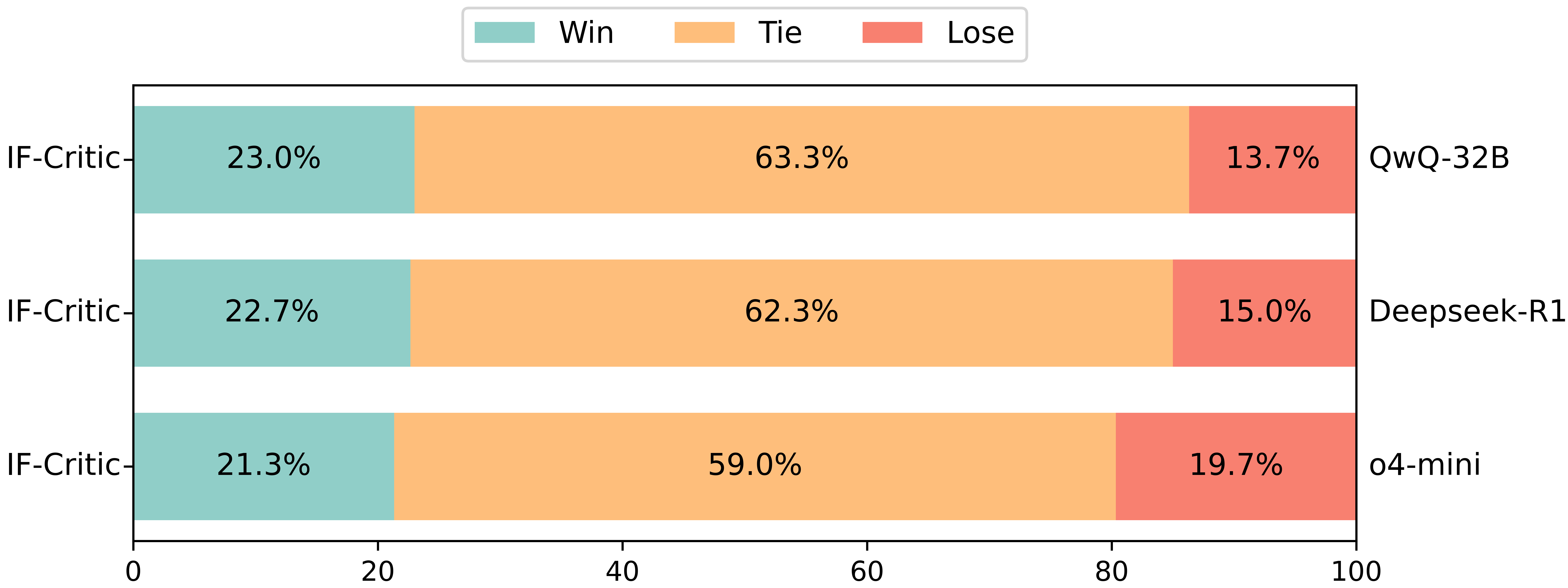}
  \caption{Explanation quality evaluation results. The percentages indicate the preference between \model and other evaluation models via human annotation.}
  \label{fig:humaneval}
\end{figure}

\definecolor{c2}{RGB}{223,246,245}

\begin{table*} [!t]
\centering
\resizebox{\textwidth}{!} {
\small
\begin{tabular}{lclc|ccc|ccc|ccc}
\toprule
\multirow{2}{*}{\textbf{Policy Model}} & \multirow{2}{*}{\textbf{Method}} & \multirow{2}{*}{\textbf{Reward Model / Critic}} & \multirow{1}{*}{\textbf{Relative}} & \multicolumn{3}{c|}{\textbf{Multi-IF}} & \multicolumn{3}{c|}{\textbf{CFBench}} & \multicolumn{3}{c}{\textbf{SysBench}} \\
\cmidrule{5-13}
& & & \multirow{1}{*}{\textbf{Time}} & Turn 1 & Turn 2 & Turn 3  & CSR   & ISR & PSR & CSR   & ISR & SSR   \\
\midrule
Qwen2.5-7B-Instruct & \multirow{2}{*}{-} & - & -  & 76.14  & 61.65 & 51.41 & 0.79 & 0.46 & 0.56 & 60.33 & 35.72 & 19.10 \\
Qwen2.5-32B-Instruct & & - & -  & 83.02  & 72.43 & 63.12 & 0.84 & 0.55 & 0.65 & 81.76 & 66.44 & 44.83 \\
\midrule
\multirow{6}{*}{Qwen2.5-7B-Instruct} & \multirow{3}{*}{DPO} & Skywork-Reward-V2-Llama-3.1-8B-40M & 0.79 & 77.86  &  63.12 & 53.15 & 0.82 & \textbf{0.53} & \textbf{0.63} & 61.59 & 38.60 & 23.60 \\
 & & QwQ-32B & 13.4 & 80.44  & 66.05  & 56.82 & 0.82 & 0.50 & 0.61 & 64.85 & 41.32 & 24.23 \\
 & & \model-14B (Ours) & 1.00 & \textbf{81.25} & \textbf{67.51}  & \textbf{57.73}  & \textbf{0.83} & 
 0.52 & \textbf{0.63} & \textbf{67.40} & \textbf{44.84} & \textbf{28.71} \\
\cmidrule{2-13}
 & \multirow{3}{*}{GRPO} & Skywork-Reward-V2-Llama-3.1-8B-40M & 1.04 & 75.28  & 61.12  & 51.90 & 0.81 & 0.50 & 0.60 & 65.82 & 44.48 & 25.79 \\
& & QwQ-32B & 3.08 & 78.59  & 63.51  & 52.60 & 0.83 & 0.54 & 0.64 & 76.25 & 59.28 & 37.58 \\
& & \model-14B (Ours) & 1.00 & \textbf{81.87}  & \textbf{67.36}  & \textbf{58.76} &  \textbf{0.85} & \textbf{0.59} & \textbf{0.69} & \textbf{81.19} & \textbf{66.80} & \textbf{44.44} \\
\midrule
\midrule
LLama3.1-8B-Instruct & \multirow{2}{*}{-} & - & -  & 73.13  & 64.27 & 56.84 & 0.70 & 0.32 & 0.43 & 60.14 & 37.32 & 19.38 \\
LLama3.1-70B-Instruct & & - & -  & 83.06 & 74.25 & 66.78 & 0.82 & 0.52 & 0.62 & 78.64 & 62.52 & 39.49 \\
\midrule
\multirow{6}{*}{LLama3.1-8B-Instruct} & \multirow{3}{*}{DPO} & Skywork-Reward-V2-Llama-3.1-8B-40M &  0.79 &  76.71 & 69.04 & 60.59 & 0.75 & 0.38 & 0.50 & 67.88 & 47.53 & \textbf{28.22} \\
 & & QwQ-32B & 12.8 & 79.72 & 71.94 & 62.97 & \textbf{0.76} & \textbf{0.41} & 0.52 & 68.36 & 46.97 & 27.53 \\
 & & \model-14B (Ours) & 1.00 & \textbf{80.75}  & \textbf{72.52} & \textbf{64.13} & \textbf{0.76} & \textbf{0.41} & \textbf{0.53} & \textbf{69.67} & \textbf{49.04} & 28.15 \\
\cmidrule{2-13}
 & \multirow{3}{*}{GRPO} & Skywork-Reward-V2-Llama-3.1-8B-40M & 1.33 & 75.57  & 67.88  & 58.59 & 0.77 & 0.42 & 0.53 & 76.99 & 59.96 & 37.42 \\
& & QwQ-32B & 3.10 & 80.67 & 72.49  & 64.40  & 0.77 & 0.44 & 0.53 & 78.22 & 62.32 & 40.79 \\
& & \model-14B (Ours) & 1.00 & \textbf{83.83} & \textbf{73.32} & \textbf{65.05} & \textbf{0.81} & \textbf{0.50} & \textbf{0.59} & \textbf{79.90} & \textbf{66.40} & \textbf{44.27} \\
\bottomrule
\end{tabular}
}
\caption{Experimental results (\%) on Multi-IF, CFBench, and SysBench. 
The best performance among different critics is \textbf{bold}. 
\textbf{Relative Time} refers to the relative time to \model of reward calculation in DPO and average per-step training in GRPO. 
Skywork-Reward-V2
is a reward model that generates scalar rewards, whereas QwQ-32B and \model generate natural language critiques. }
\label{tab:policy_model_performance}
\end{table*}
\paragraph{Analysis on Explanation Quality.} 
To assess the quality of the explanations generated by \model, we randomly sample 300 constraints from the first three evaluation benchmarks and collect corresponding generated explanations from \model, QwQ-32B, Deepseek-R1, and o4-mini.
For each pair of explanations (one from \model and the other from a baseline evaluation model, given the same evaluation input), we ask human annotators to judge which explanation is better (i.e., win, lose, or tie) in terms of correctness, informativeness, and helpfulness.
The priority of these aspects follows the above order. 
As shown in Figure \ref{fig:humaneval}, \model surpasses QwQ-32B and Deepseek-R1 in win rates by 9.3\% and 7.7\%, respectively, and performs on par with o4-mini.
Many results are ties because all four evaluators are strong and correct on many constraints. 
This demonstrates that \model can generate high-quality explanations for instruction-following evaluation.

\subsection{Instruction-Following Optimization}
\label{instruction_following_optimization}
\paragraph{Training Datasets \& Settings.}
To collect diverse and high-quality complex instructions for model training, we first curate seed instructions from real-world application scenarios. 
These scenarios are distinct from the training data of \model\ and generally exhibit lower instruction complexity. 
Stratified sampling based on task categories is adopted to enhance diversity.  
Subsequently, we employ Deepseek-R1 to introduce constraints to these instructions based on a comprehensive constraint taxonomy that encompasses 6 primary constraint categories and 4 constraint composition types. 
Deepseek-R1 is also used to validate the generated instructions, ensuring they are reasonable, unambiguous, and logically consistent. 
In total, we obtain 18.5k high-quality complex instructions for training.
Further details are in Appendix \ref{app:instruction_following_optimization}.

We choose Qwen2.5-7B-Instruct and Llama-3.1-8B-Instruct as the initial policy models for DPO and GRPO training.
Three evaluation models are used to provide reward signals: 
\model, the state-of-the-art reward model Skywork-Reward-V2-Llama-3.1-8B-40M, and the powerful open-source general LLM QwQ-32B. 
For DPO training, we sample 10 responses per instruction, with a temperature of 1.0 and a top-p value of 0.9. 
The responses with the highest and lowest rewards are used to construct preference pairs. 
For GRPO training, we perform 32 rollouts per instruction. 
More training details are in Appendix \ref{app:instruction_following_optimization}.

\paragraph{Evaluation Benchmarks.}
We evaluate our models on three popular instruction-following
benchmarks: \textbf{Multi-IF} \cite{he2024multi},
which assesses multi-turn and multilingual hard constraints following; \textbf{CFBench} \cite{zhang2024cfbench}, which covers a comprehensive range of constraint types; and \textbf{Sysbench} \cite{qin2025sysbench}, which focuses on the following of system prompts. GPT-4o-2024-11-20 is used as the evaluation model for CFBench and SysBench.
For \textbf{Multi-IF}, we follow RPO \cite{huang2025reverse} and report the average performance across the English and Chinese subsets.
\paragraph{Main Results.}
The main results are presented in Table \ref{tab:policy_model_performance}. 
\textbf{Firstly}, optimization with \model brings substantially superior performance gains across various benchmarks compared to all strong baselines, substantiating its advanced capacity of instruction-following evaluation. 
Notably, LLMs after GRPO training with \model even achieve comparable performance to significantly larger LLMs (32B or 70B) of the same model family.
In contrast, the general reward model, Skywork-Reward-V2, is ill-suited for instruction-following optimization, yielding only marginal improvements.
\textbf{Secondly}, compared to the powerful LLM critic, QwQ-32B, \model only incurs less than a third of computational overhead, highlighting its practicality. 
\textbf{Finally}, 
GRPO proves more effective for instruction-following tasks compared to DPO, delivering greater performance gains. 
The reward curves during GRPO training are provided in the Appendix \ref{app:reward_curves}.
We also validate that optimization with \model does not compromise the general performance of LLMs in the Appendix \ref{app:general_preformance}.

\section{Conclusion}

Our work proposes \model, an LLM critic for fine-grained instruction-following evaluation. 
\model adopts a checklist-guided critique generation paradigm that evaluates all constraints in a checklist within one inference pass, improving efficiency and reliability. 
We collect high-quality critique training data through a multi-stage filtering mechanism and employ constraint-level preference optimization to train \model. 
Extensive experiments show that \model has stronger evaluation ability and can bring superior performance gains to LLMs in instruction-following optimization, 
compared to powerful LLM critic baselines.

\section*{Limitations}
The limitations of our work are summarized as follows:
\paragraph{Scope of Rule-Augmented Verification.}
In the critique training data construction stage of \model, we only introduce rule-augmented verification for length-related constraints, while other code-verifiable constraints (e.g., keywords or structural format) are not comprehensively covered. 
We prioritize length-related constraints because they are prevalent in real-world complex instructions, and the previous version of \model performs well on other verifiable constraints but struggles with length constraints, as observed in our primary experiments. 
Although this approach has yielded substantial performance gains, extending this approach to a broader spectrum of constraint types remains an important direction for future work.
\paragraph{Potential Evaluation Bias.} 
Similar to other model-based evaluation methods, \model may also suffer from potential evaluation biases such as self-enhancement or verbosity bias \cite{zheng2023judging}. 
These biases could harm the evaluation correctness of certain models.
Although our multi-stage critique filtering mechanism has effectively mitigated noise and bias in training critique collection, incorporating inference-time strategies (e.g., multi-agent debate) may further attenuate bias and enhance evaluation performance. 
We reserve further investigation of evaluation bias mitigation as important future work.

\section*{Ethical Considerations}
In this work, we recruit some human annotators to validate the quality of training data and the performance of \model.
The annotator pool primarily consists of college students. 
Throughout the data annotation process, we adhere to several key principles.
Firstly, all annotators are fully informed about the purpose of our study and the involved tasks. 
Secondly, all annotators received fair compensation for their contributions based on the market price. 
Finally, any harmful instructions are filtered before annotation to avoid ethical issues.

\section*{Acknowledgments}
This work was supported by the National Science Foundation for Distinguished Young Scholars (with No. 62125604) and the Natural Science Foundation of China (No. 62536008). 
We would also like to thank Zhipu AI for sponsoring the computational resources and annotation costs in this work.

\bibliography{custom}

\appendix

\begin{table*} [!t]
\centering
\resizebox{\textwidth}{!} {
\small
\begin{tabular}{ll|cc|cc|cc|cc|c}
\toprule
\multirow{3}{*}{\textbf{Model}}  & \multirow{3}{*}{\textbf{Decoding Strategy}} & \multicolumn{2}{c|}{\textbf{EvalCritic}} & \multicolumn{2}{c|}{\textbf{CFBench}} & \multicolumn{2}{c|}{\textbf{TRACE}} & \multicolumn{2}{c|}{\textbf{Multi-IF}} & \textbf{Avg.} \\
\cmidrule{3-11}
 &  & Positive   & Negative   & Positive   & Negative  &  Positive   & Negative & Positive   & Negative & Average \\
 & & F1 & F1  & F1 & F1  & F1 & F1 & F1 & F1 & F1 \\
\midrule
\multirow{2}{*}{Deepseek-R1} & Greedy Search  & 0.895   & \textbf{0.716} & 0.928   & 0.725 &  0.910 & \textbf{0.579} & \textbf{0.938} & 0.828 &  \textbf{0.815} \\
&  Default & \textbf{0.901}   & 0.702 & \textbf{0.933} & \textbf{0.730} &  \textbf{0.915} & 0.565  & 0.936 & \textbf{0.829} &  0.814 \\
\midrule
\multirow{2}{*}{QwQ-32B} & Greedy Search & \textbf{0.892}  & \textbf{0.663}  & 0.922 & \textbf{0.716}  & \textbf{0.922} & \textbf{0.570} & 0.933 & 0.793 &  \textbf{0.801} \\
& Default & 0.882 & 0.643 & \textbf{0.928} & 0.713 & 0.919 & 0.555 & \textbf{0.937} & \textbf{0.805} &  0.798 \\
\bottomrule
\end{tabular}
}
\caption{Positive F1 and Negative F1 scores of constraint-following verification on the instruction-following evaluation task under different decoding strategies.}
\label{tab:decoding_strategies}
\end{table*}

\section{Model List for Response Generation}
\label{app:model_list}
We use 15 representative LLMs for response generation, including GPT-4o \cite{openai2023gpt4}, GPT-4o-mini \cite{openai2023gpt4},
seven versions of GLM-4 \cite{glm2024chatglm}, three versions
of Qwen2.5 \cite{yang2024qwen2}, Deepseek-R1 \cite{guo2025deepseek}, Doubao1.5 \cite{doubao2025doubao}, and Claude3.7-sonnet \cite{claude}. 

\section{List for Prompt Templates}
\label{app:prompt_templates}
This section lists all the prompt templates applied
throughout this work, including the prompt for \textbf{scoring the quality of user instruction} in Table \ref{tab:scoring_prompt}, the prompt for \textbf{constraint checklist generation} in Table \ref{tab:checklist_generation}, the prompt for \textbf{checklist-guided critique generation} in Table \ref{tab:critique_generation}, the prompt for \textbf{cross-model verification} in Table \ref{tab:cross_validation}, and the prompt for \textbf{rule-augmented verification} in Table \ref{tab:length_constraint_identification} and \ref{tab:rule_revision}.

\section{List for Baseline Evaluation Models}
\label{app:baseline_models}
For general baselines, we adapt 3 advanced proprietary LLMs, GPT-4.1 \cite{openai2023gpt4}, o4-mini \cite{jaech2024openai}, and Gemini-3-Pro \cite{team2023gemini}, as well as 5 strong open-source LLMs, GLM-4-32B-0414 \cite{glm2024chatglm}, LLama-3.3-70B-Instruct \cite{grattafiori2024llama}, Qwen2.5-72B-Instruct \cite{yang2024qwen2}, QwQ-32B \cite{qwq32b}, and Deepseek-R1 \cite{guo2025deepseek}.
For evaluation-specific baselines, we select 4 powerful discriminative or generative reward models, Skywork-Reward-V2-Llama-3.1-8B-40M \cite{liu2025skywork}, Prometheus-BGB-8x7B \cite{kim-etal-2024-prometheus}, RM-R1-Deepseek-Distilled-Qwen-32B \cite{chen2025rm}, and RRM-32B \cite{guo2025reward}.

\section{Implementation Details of \model}
\label{app:training_details}
We use the Llama-Factory \cite{zheng-etal-2024-llamafactory} framework and the Zero Redundancy Optimizer (ZeRO) \cite{rajbhandari2020zero} stage 3 from the Deepspeed library \cite{rasley2020deepspeed} to train \model. 
All experiments are conducted on 8 H800 GPUs.
We utilize the AdamW \cite{kingma2014adam} optimizer with a weight decay of 0.1, and set the maximum sequence length to 8192 tokens during all the experiments.

During the supervised fine-tuning, the peak learning rate is set to 5e-6 with a 10\% warmup ratio and a linear scheduler. 
The batch size is 16, and training is conducted for 2 epochs.
During the constraint-level preference-optimization,
the peak learning rate is set to 1e-6 with a 10\% warmup ratio and a cosine scheduler. 
The batch size is 128.
Training utilizes a sigmoid loss function with a beta value of 0.1 and spans 1 epoch.
Following previous works \cite{hou2024chatglm}, an additional SFT loss is added to the chosen critique with a weight of 0.1.
The overall API cost for critique training data collection is approximately 4000\$.

\section{Experimental Results on Other Decoding Strategies}
\label{app:decoding_strategies}

In our preliminary experiment, we have compared the performance of greedy search decoding against sampling-based decoding strategies. 
Specifically, we conduct additional experiments on DeepSeek-R1 and QwQ-32B using their default decoding strategies (a temperature of 1.0 and a top-p value of 1.0 for DeepSeek-R1; a temperature of 0.6, a top-p value of 0.95, and a top-k value of 20 for QwQ-32B). 
The F1 scores of constraint-following verification using the checklist-level prompting strategy are presented in Table \ref{tab:decoding_strategies}. 
We find that greedy decoding performs almost identically to their default decoding strategies and does not hurt model performance. 
Consequently, we use greedy decoding to ensure reproducibility.

\begin{table} [!t]
\centering
\resizebox{\linewidth}{!} {
\begin{tabular}{l|c|c|c|c}
\toprule
\textbf{Base Model} & \textbf{EvalCritic} & \textbf{CFBench}  & \textbf{TRACE} & \textbf{Multi-IF} \\
\midrule
Qwen2.5-14B-Insturct & 0.861 & 0.863   &  \textbf{0.840} & \textbf{0.895} \\
\midrule
Qwen-2.5-7B-Instruct & 0.822 & 0.798   &  0.806 & 0.813 \\
Qwen-3-8B & 0.855 &  0.823 & 0.820 & 0.848 \\
GLM-4-9B-0414 & 0.797 & 0.810  &  0.801 & 0.810 \\
Qwen-3-14B & 0.868 & \textbf{0.863}   &  0.828 & 0.860 \\
Qwen-2.5-32B-Instruct & \textbf{0.879} & 0.854   &  0.833 & 0.867 \\
\bottomrule
\end{tabular}
}
\caption{The average of Positive F1 and Negative F1 scores of \model under different base models. Qwen-2.5-14B-Instruct is the base model we used for the main experiments.}
\label{tab:base_model_ablation}
\end{table}

\section{Experimental Results on Other Base Models}
\label{app:base_model}

In our primary experiment, we experiment with various base models for 
\model, and the results are presented in Table \ref{tab:base_model_ablation}. 
We observe that LLMs with approximately 8B parameters are still insufficient to fully acquire critique generation capabilities, exhibiting a notable performance gap to larger LLMs. 
In contrast, LLMs with approximately 14B parameters are generally adequate for learning critique generation, and further increasing model scale yields negligible performance improvement.
Considering the trade-off between performance and efficiency, we ultimately choose Qwen-2.5-14B-Instruct as the base model for our main experiments.

\section{Details of Instruction-Following Optimization}
\label{app:instruction_following_optimization}
\paragraph{Training Instructions Collection.} To ensure the diversity and complexity of constraints, we first construct a comprehensive and hierarchical constraint taxonomy, which consists of 6 primary categories (i.e., format constraints, content constraints, behavioral constraints, style constraints, role constraints, and environmental constraints) and 54 secondary categories. 
During constraint synthesis, we randomly sample 2 to 5 secondary constraint categories and require Deepseek-R1 to introduce several corresponding constraints into the seed instructions, thereby increasing their complexity.
In contrast to previous work that only considers independent and atomic constraints \cite{jiang2023followbench, ren-etal-2025-step, peng-etal-2025-verif}, we further ask Deepseek-R1 to compose multiple constraints according to the composition types proposed in ComlexBench \cite{wen2024benchmarking}, including: (1) \textit{And}: Coordination between different constraints, (2) \textit{Chain}: Sequential completion of constraints, (3) \textit{Selection}: Conditional selection of constraints, and (4) The nested structure of the above composition types.
In this way, our constraint synthesis can more comprehensively encompass the complex instruction types in real-world scenarios.

In this scenario, it is notable that most of the hard constraints cannot be reliably verified using automatic verification codes alone, such as constraints targeting for specific segments of the response (e.g., \textit{Generate 10 titles, each no more than 8 words}) and the composition of soft and hard constraints (e.g., \textit{Bold all adjectives that express feelings, such as \textbf{happy}}), as it is challenging for verification code to accurately extract the corresponding segments of the response that should follow these constraints.
We provide some examples of the seed instructions and the corresponding final constructed complex instructions in Table \ref{tab:training_data_prompts}.

\begin{table*} [!t]
\centering
\resizebox{\textwidth}{!} {
\small
\begin{tabular}{lcl|c|c|c|c|l}
\toprule
\textbf{Policy Model} & \textbf{Method} & \textbf{Reward Model / Critic} & \textbf{AlignBench} & \textbf{MMLU-Pro} & \textbf{Omni-MATH} & \textbf{HumanEval} & \textbf{Avgerage} \\
\midrule
\multirow{8}{*}{Qwen2.5-7B-Instruct} & - & - &    58.61 & 57.32 & 32.20 & 79.27 & 56.85  \\
\cmidrule{2-8}
& \multirow{3}{*}{DPO} & Skywork-Reward-V2 &  61.36  & 57.41 & 31.95 & 75.00 &  56.43{\tiny \textbf{\textcolor{purple}{(-0.42)}}} \\
 & & QwQ-32B &  60.81  & 57.14 & 32.05 & 77.44 & 56.86{\tiny \textbf{\textcolor{teal}{(+0.01)}}} \\
 & & \model-14B (Ours) &  61.16   & 57.20 & 32.10 & 77.44 & 56.98{\tiny \textbf{\textcolor{teal}{(+0.13)}}} \\
\cmidrule{2-8}
 & \multirow{3}{*}{GRPO} & Skywork-Reward-V2 &  61.49   & 57.43 & 31.18 & 74.39 & 56.12{\tiny \textbf{\textcolor{purple}{(-0.73)}}}  \\
& & QwQ-32B &  62.64  & 57.51 & 32.69 & 73.78 & 56.66{\tiny \textbf{\textcolor{purple}{(-0.19)}}} \\
& & \model-14B (Ours) &  64.83  & 56.57 & 32.11 & 76.83 & 57.59{\tiny \textbf{\textcolor{teal}{(+0.74)}}} \\
\midrule
\midrule
\multirow{8}{*}{LLama3.1-8B-Instruct} & - &  - &    40.66 & 44.17 & 16.05 & 64.02 & 41.23 \\
\cmidrule{2-8}
 & \multirow{3}{*}{DPO} & Skywork-Reward-V2 &   44.93   & 44.75 & 15.44 & 63.41 & 42.13{\tiny\textbf{\textcolor{teal}{(+0.90)}}} \\
 & & QwQ-32B &  41.23  & 44.61 & 16.54 & 65.85 & 42.06{\tiny\textbf{\textcolor{teal}{(+0.83)}}} \\
 & & \model-14B (Ours) &  45.76 & 44.11 & 16.02 & 63.41 & 42.33{\tiny\textbf{\textcolor{teal}{(+1.10)}}} \\
\cmidrule{2-8}
& \multirow{3}{*}{GRPO} & Skywork-Reward-V2 &    46.97  & 45.65 & 17.38 & 64.02 & 43.51{\tiny\textbf{\textcolor{teal}{(+2.28)}}} \\
&  & QwQ-32B  & 45.95 & 44.76 & 17.25 & 60.37 & 42.08{\tiny \textbf{\textcolor{teal}{(+0.85)}}} \\
& & \model-14B (Ours) &  46.26  & 45.52 & 17.75 & 60.98 & 41.98{\tiny \textbf{\textcolor{teal}{(+1.40)}}} \\
\bottomrule
\end{tabular}
}
\caption{Experimental results (\%) on various general benchmarks. The original results of AlignBench range from 1 to 10; we convert the scale to 10-100 for ease of calculating average performance. Skywork-Reward-V2 refers to Skywork-Reward-V2-Llama-3.1-8B-40M. }
\label{tab:general_performance}
\end{table*}
\paragraph{Training Configuration.} For DPO training, we use the Llama-Factory \cite{zheng-etal-2024-llamafactory} framework, and all other settings are the same as the preference optimization for \model. 
During the reward calculation, 
\model and QwQ-32B utilize the vllm \cite{kwon2023efficient} inference framework, while Skywork-Reward-V2-Llama-3.1-8B-40M is deployed by sglang\footnote{https://github.com/sgl-project/sglang} (since this reward model is not compatible with the vllm framework currently).
The reward calculation and DPO training are both conducted on 8 H800 GPUs.

For GRPO training, we use the VeRL \cite{sheng2025hybridflow} framework.
The learning rate is set to 2e-6, with the maximum sequence length for both prompts and responses set to 4096 tokens. 
The train-batch size and PPO mini-batch size are each set to 32. 
For rollout responses generation, we employ the vllm \cite{kwon2023efficient} framework, utilizing 60\% of the available GPU memory,  with a temperature and top-p value of 1.0.
To encourage stable training, we also incorporate KL divergence regularization with a coefficient of 0.001, using the low-variance KL implementation. 
We save checkpoints every 30 steps during training. 
Following TULU 3 \cite{lambert2024t} and VerIF \cite{peng-etal-2025-verif}, we use IFEval \cite{zhou2023instruction} as the validation set to select the best checkpoint.
All experiments are conducted on 24 H800 GPUs, with 8 GPUs dedicated to model training and the remaining 16 GPUs allocated for API deployment of the instruction-following critic or reward model.
The deployment framework is identical to the inference framework described in the above DPO training.

\section{Reward Curves During GRPO Training}
\label{app:reward_curves}
\begin{figure*}[!t]
  \centering
  \includegraphics[width=1.0\textwidth]{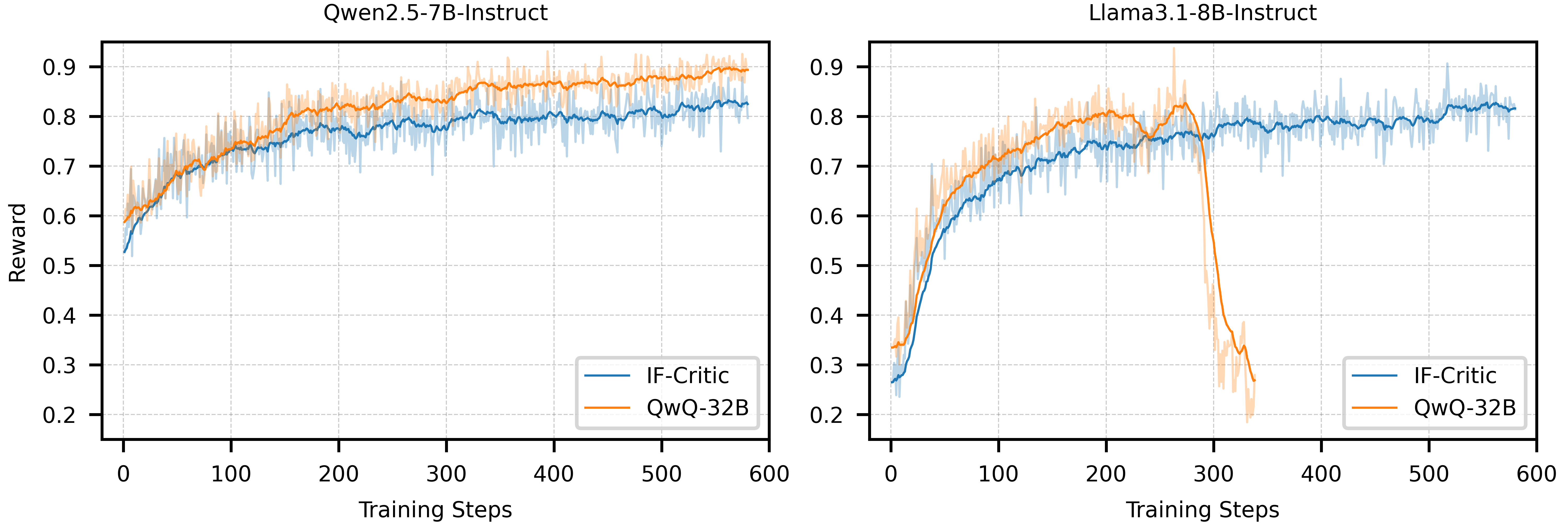}
  \caption{Reward curves during GRPO training when \model and QwQ-32B are employed as the LLM critics. For LLama-3.1-8B-Instruct, training with QwQ-32B results in a model collapse after 300 steps, with the model tending to generate extensive repetitive and meaningless content. Due to efficiency considerations, we terminate further training and calculate the average per-step training time for all critics and reward models based on the first 300 steps.}
  \label{fig:critic_reward}
\end{figure*}

\begin{figure*}[!t]
  \centering
  \includegraphics[width=1.0\textwidth]{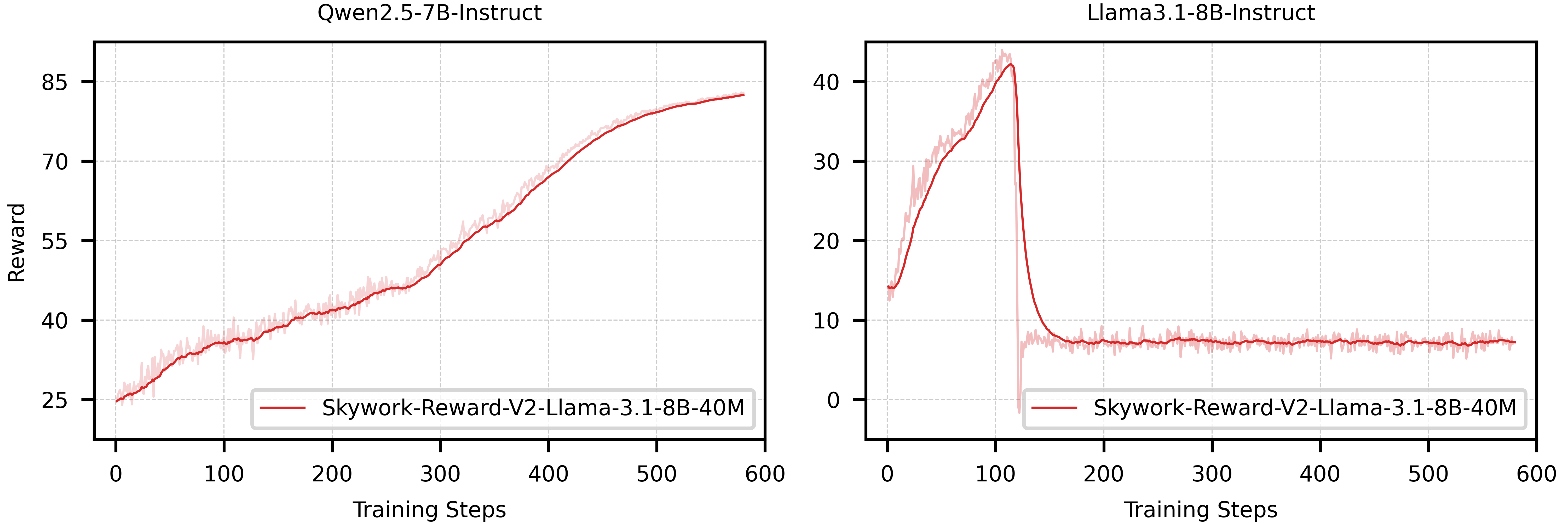}
  \caption{Reward curves during GRPO training when Skywork-Reward-V2-Llama-3.1-8B-40M is employed as the reward model.}
  \label{fig:skywork_reward}
\end{figure*}

We provide the reward curves during GRPO training when \model and QwQ-32B are employed as the LLM critic in Figure \ref{fig:critic_reward}, and Skywork-Reward-V2-Llama-3.1-8B-40M as the reward model in Figure \ref{fig:skywork_reward}. 
IF-Critic has a stronger error-detection capability in instruction following, which leads to it generally assigning lower rewards to policy models than QwQ-32B.
We also observe that the rewards provided by Skywork-Reward-V2-Llama-3.1-8B-40M may have potential length bias. 
Using it as a reward model for GRPO training significantly increases the average response length of the policy model, resulting in a longer average per-step training time compared to \model.

\section{Experimental Results on General Benchmarks}
\label{app:general_preformance}
We provide the performance of LLMs after instruction-following optimization in four general benchmarks: AlignBench \cite{liu-etal-2024-alignbench}, MMLU-Pro \cite{wang2024mmlupro}, Omni-MATH \cite{gao2025omnimath}, and HumanEval \cite{chen2021evaluating} in Table \ref{tab:general_performance}. 
GPT-4o-2024-11-20 is used as the evaluation model for AlignBench, and we use OmniJudge\footnote{https://huggingface.co/KbsdJames/Omni-Judge} for the evaluation of Omni-MATH.
The results reveal that conducting instruction-following optimization with \model does not harm the general performance of LLMs.

\section{Human Annotation Guideline for Constraint Verification}
The human annotation guideline for constraint verification in the section \textbf{Instruction-Following Evaluation} is shown in Table \ref{tab:human_annotation}. 
Given an instruction and a corresponding model response, as well as the constraint checklist for the instruction, human annotators are instructed to judge whether the model response follows each constraint in the checklist.

\begin{table*} [t]
\centering
\small
\setlength{\tabcolsep}{1.6mm}{
% \begin{tabular}{l >{\raggedright\arraybackslash}m{0.85\textwidth}}
\begin{tabular}{p{\dimexpr 0.16\linewidth-2\tabcolsep\relax}|p{\dimexpr 0.84\linewidth-2\tabcolsep\relax}}
\toprule
\textbf{Type} & \textbf{Prompt} \\
\midrule
Seed Instruction & 
You’re an expert in artificial intelligence training. How is a machine learning model built? Please explain in a clear and easy-to-understand tone. \\
\midrule
Final Instruction & 
You’re an expert in artificial intelligence training. Please explain the process of building a machine learning model in five stages, following a “general-to-specific-to-general” structure. Ensure the following requirements are satisfied:
\newline
1. Use the symbols \ding{182}\ding{183}\ding{184} to notate each sub-step within each stage.
\newline
2. Do not include any mathematical formulas or code examples.
\newline
3. Integrate real-life kitchen or cooking analogies within each stage.
\newline
4. Finally, use $\triangleright$ to highlight three common mistakes people make when building a machine learning model.
\newline
5. Write in the second-person perspective, ensure no technical terms exceed middle school math knowledge, and make sure all analogies involve kitchen or cooking elements.
 \\
\midrule
\midrule
Seed Instruction & 
Translate the following English words into Chinese:
\newline
attachment, surpasses, dissent, retention, unobtrusive,  glazed, entails, confederates, skeptical, perceive, enormity, fanatical \\
\midrule
Final Instruction & 
Please translate the following English words into Chinese, strictly following these requirements:
\newline
1. Precede each translated Chinese word with a numbered bullet point (e.g., \ding{172}, \ding{173}, \ding{174}).
\newline
2. Indicate the part of speech of the original English word (e.g., noun/verb/adjective) in parentheses after each translation.
\newline
3. Provide a Chinese synonym after each Chinese translation word, separated by a dash.
\newline
\newline
Words to be translated:
\newline
attachment, surpasses, dissent, retention, unobtrusive,  glazed, entails, confederates, skeptical, perceive, enormity, fanatical 
 \\
\midrule
\midrule
Seed Instruction & 
Write a conversation between a Chinese international student and a Japanese student who decide to try a new restaurant together. While they're having their meal, one of them suggests that they go to see a movie afterwards. \\
\midrule
Final Instruction & 
Please write a conversation between Xiaolin, a Chinese international student, and Yamada, a Japanese student. 
In the conversation, they decide to try a new restaurant together. While they're having their meal, one of them suggests that they go to see a movie afterwards. Ensure the following requirements are satisfied:
\newline
1. The conversation should contain three stages: (1) \textbf{Stage 1: Arranging to dine at the restaurant} (at least 2 turns), (2) \textbf{Stage 2: Discussing the food during the meal} (at least 3 turns), (3) \textbf{Stage 3: Proposing to watch a movie} (at least 3 turns). Separate each stage of the conversation with two blank lines.
\newline
2. If the restaurant is a Chinese restaurant, the suggested movie genre should be kung fu films. If the restaurant is a sushi restaurant, the suggested movie genre should be animated films.
\newline
3. Xiaolin should naturally mention Mid-Autumn Festival customs in the conversation while Yamada should incorporate elements of the Obon Festival. Please use specific festival details (e.g., types of mooncakes, or forms of Bon Odori).
\newline
4. Please write the conversation in Chinese. Indicate the speaker (Xiaolin: / Yamada:) at the start of each turn. The entire conversation should be no more than 10 turns.
\newline
\newline
Please first determine the restaurant type (Chinese restaurant or sushi restaurant), and then write the conversation according to the corresponding scene.
 \\
\bottomrule
\end{tabular}
}
\caption{Some examples of the seed instructions and the corresponding final constructed complex instruction in the instruction-following optimization.}
\label{tab:training_data_prompts}
\end{table*}
\begin{table*} [!t]
\centering
\setlength{\tabcolsep}{1.6mm}{
\begin{tabular}{p{\dimexpr \linewidth-2\tabcolsep\relax}}
\toprule
You are an expert specializing in evaluating the quality of user instructions. 
Please evaluate the quality of the user instruction based on the following criteria.
\newline
\newline
\#\# Evaluation Criteria:
\newline
1. \textbf{Low Quality}: The user instruction contains significant issues such as unclear, incomplete, or ambiguous information, making it impossible to determine the intent behind the instruction.
\newline
2. \textbf{Medium Quality}: The user instruction may include some unclear, incomplete, or ambiguous elements; however, the overall intent can still be inferred.
\newline
3. \textbf{High Quality}: The user instruction is clear, complete, and unambiguous, allowing the intent to be easily and definitively understood.
\newline
\newline
\#\# Note:
\newline
1. You do not need to answer the user instruction, but only need to output the evaluation result.
\newline
2. If the user instruction requires additional retrieval or the use of tools to obtain an answer, it should be evaluated as \textbf{Medium Quality}.
\newline
\newline
Here are some examples and the user instruction to be evaluated:
\newline
\textbf{[The Start of Examples]}
\newline
\textcolor{red}{\{in\_context\_examples\}}
\newline
\textbf{[The End of Examples]}
\newline
\newline
\textbf{[The Start of User Instruction]}
\newline
\textcolor{red}{\{instruction\}}
\newline
\textbf{[The End of User Instruction]}
\newline
\newline
\#\# Output Format
\newline
"""
\newline
Prompt Quality: \textbf{Low Quality} / \textbf{Medium Quality} / \textbf{High Quality}
\newline
"""
\newline
\newline
\#\# Please directly output the evaluation result:
\\
\bottomrule
\end{tabular}
}
\caption{The prompt template for scoring the quality of user instructions. }
\label{tab:scoring_prompt}
\end{table*}
\begin{table*} [!t]
\centering
\setlength{\tabcolsep}{1.6mm}{
\begin{tabular}{p{\dimexpr \linewidth-2\tabcolsep\relax}}
\toprule
You are an information extraction expert specializing in identifying all constraints within instructions.
\newline
\newline
We will provide you with a user instruction. Your task is to extract all constraints present in that instruction.
\textbf{Constraints} refer to all instructional content,  excluding auxiliary information such as \textbf{background knowledge}, \textbf{text materials}, and \textbf{in-context examples}. 
This includes, but is not limited to, descriptions of basic tasks to be completed, and specific requirements on output content, style, and format.
Output all constraints using this format exactly:
\newline
\newline
"""
\newline
\textbf{[The Start of Constraint 1]}
\newline
\textbf{Constraint:} ... (Please generate a specific constraint of the instruction. It must be complete and detailed, with no information omitted or altered in any way)
\newline
\textbf{[The End of Constraint 1]}
\newline
\newline
\textbf{[The Start of Constraint 2]}
\newline
\textbf{Constraint:} ... (Please generate a specific constraint of the instruction. It must be complete and detailed, with no information omitted or altered in any way)
\newline
\textbf{[The End of Constraint 2]}
\newline
\newline
…
\newline
"""
\newline
\newline
Remember the following points:
\newline
(1) Each constraint you output must be a verbatim excerpt from the given instruction. You must not fabricate, paraphrase, or add any content.
\newline
(2) You must extract and output all constraints in the order in which they appear in the instruction, without omitting any constraints or inventing constraints not present in the instruction. To reiterate: \textbf{Constraints} refer to all instructional content, excluding auxiliary information such as \textbf{background knowledge}, \textbf{text materials}, and \textbf{in-context examples}. 
\newline
(3) Do not output duplicate constraints. Each portion of the given instruction should appear in at most one constraint, and must not be repeated across multiple constraints.
\newline
(4) Each constraint should be atomic and independent, without inclusion or dependency relationships. At the same time, ensure appropriate granularity of constraints: neither too fine nor too coarse. Each constraint should have complete semantics and be understandable on its own, without reference to other constraints. At the same time, relatively independent constraints must not be merged into a single constraint.
\newline
\newline
Here are the user instruction:
\newline
\textbf{[The Start of User Instruction]}
\newline
\textcolor{red}{\{instruction\}}
\newline
\textbf{[The End of User Instruction]}
\newline
\\
\bottomrule
\end{tabular}
}
\caption{The prompt template for constraint checklist generation. }
\label{tab:checklist_generation}
\end{table*}
\begin{table*} [!t]
\small
\centering
\setlength{\tabcolsep}{1.6mm}{
\begin{tabular}{p{\dimexpr \linewidth-2\tabcolsep\relax}}
\toprule
You are a fair judge, specializing in evaluating the quality of an AI assistant's responses to user instructions.
\newline
\newline
We will provide you with a user instruction, a constraint checklist for the user instruction, and an AI assistant's response.
Please think carefully and provide a detailed analysis of whether the AI assistant's response follows each item on the constraint checklist.
You need to analyze and assess every constraint in the checklist, and do not need to analyze anything that is not listed in the checklist.
You must output your explanation and judgment for each constraint, strictly following the format below:
\newline
\newline
"""
\newline
\textbf{[The Start of Constraint 1]}
\newline
\textbf{Constraint:} ... (Directly output the first constraint from the checklist here with no modifications)
\newline
\textbf{Explanation:} ... (Provide a thorough, step-by-step analysis of whether the AI assistant's response follows this constraint,  referencing specific details of the response)
\newline
\textbf{Judgment:} [[The AI assistant’s response follows this constraint]] or [[The AI assistant’s response does not follow this constraint]] 
\newline
\textbf{[The End of Constraint 1]}
\newline
\newline
\textbf{[The Start of Constraint 2]}
\newline
\textbf{Constraint:} ... (Directly output the second constraint from the checklist here with no modifications)
\newline
\textbf{Explanation:} ... (Provide a thorough, step-by-step analysis of whether the AI assistant's response follows this constraint,  referencing specific details of the response)
\newline
\textbf{Judgment:} [[The AI assistant’s response follows this constraint]] or [[The AI assistant’s response does not follow this constraint]] 
\newline
\textbf{[The End of Constraint 2]}
\newline
\newline
…
\newline
"""
\newline
\newline
Remember the following points:
\newline
(1) You should only focus on whether the response follows the items in the checklist, not on the overall quality of the response. Other shortcomings in the response (such as not fulfilling a constraint to a high standard) should not affect your judgment unless the constraint is not followed.
\newline
(2) Your judgment should be as strict as possible. You should output "[[The AI assistant’s response follows this constraint]]" only if the response completely follows every part of the constraint as stated, without any omission or mistake. 
\newline
(3) When analyzing a given constraint, you should not take into account whether other constraints are followed.
\newline
\newline
Here are the user instruction, constraint checklist, and AI assistant's response:
\newline
\textbf{[The Start of User Instruction]}
\newline
\textcolor{red}{\{instruction\}}
\newline
\textbf{[The End of User Instruction]}
\newline

\textbf{[The Start of Constraint Checklist]}
\newline
\textcolor{red}{\{checklist\}}
\newline
\textbf{[The End of Constraint Checklist]}
\newline

\textbf{[The Start of AI Assistant's Response]}
\newline
\textcolor{red}{\{response\}}
\newline
\textbf{[The End of AI Assistant's Response]}
\\
\bottomrule
\end{tabular}
}
\caption{The prompt template for checklist-guided critique generation. }
\label{tab:critique_generation}
\end{table*}
\begin{table*} [t]
\centering
\small
\setlength{\tabcolsep}{1.6mm}{
\begin{tabular}{p{\dimexpr 0.12\linewidth-2\tabcolsep\relax}|p{\dimexpr 0.88\linewidth-2\tabcolsep\relax}}
\toprule
\textbf{Template} & \textbf{Prompt} \\
\midrule
Correctness \newline Verification & 
You are a fair judge, specializing in evaluating the quality of an AI assistant’s responses to user instructions.
\newline
\newline
We will provide you with a user instruction, an AI assistant's response, a specific constraint from the user instruction, and a critique of the AI's response regarding whether the response follows this constraint from the user instruction. The critique comprises two parts: "Explanation" and "Judgment". Your task is to carefully analyze this critique to judge whether it is reasonable and correct. After your analysis, output either "[[The given critique is correct]]" or "[[The given critique is not correct]]" as your final conclusion.
\newline
\newline
A correct critique must simultaneously satisfy all of the following three principles:
\newline
(1) Only focus on whether the response follows the constraint, not on the overall quality of the response. Other shortcomings in the response (such as not fulfilling the constraint to a high standard) should not affect the judgment unless the constraint is not followed.
\newline
(2) Be as strict as possible. Output "[[The AI assistant’s response follows this requirement]]" only if the response completely
follows every part of the requirement as stated, without any omission or mistake.
\newline
(3) Independently analyzes the given constraint, and does not take into account whether other constraints are followed.
\newline 
\newline
Here are the user instruction, the constraint, the AI assistant's response, and the critique to be judged:
\newline
\textbf{[The Start of User Instruction]}
\newline
\textcolor{red}{\{instruction\}} 
\newline
\textbf{[The End of User Instruction]}
\newline
\newline
\textbf{[The Start of Constraint]}
\newline
\textcolor{red}{\{constraint\}}
\newline
\textbf{[The End of Constraint]}
\newline
\newline
\textbf{[The Start of AI Assistant's Response]}
\newline
\textcolor{red}{\{response\}}
\newline
\textbf{[The End of AI Assistant's Response]}
\newline
\newline
\textbf{[The Start of Critique]}
\newline
\textcolor{red}{\{critique\}}
\newline
\textbf{[The End of Critique]} \\
\midrule
Consistency \newline  Verification & 
You are a fair judge, specializing in evaluating the quality of an AI assistant’s responses to user instructions.
\newline
\newline
We will provide you with a constraint, a critique that comprises an explanation and a judgment generated by an evaluator, whose main purpose is to analyze whether an  AI assistant's response follows the constraint. You do not need to consider the content of the constraint or whether the critique is correct. You only need to judge whether the explanation and judgment within the critique are logically consistent, specifically:
\newline
(1) If the explanation section determines that the response follows the constraint, the judgment should be: "[[The AI assistant's response follows the constraint]]".
\newline
(2) If the explanation section identifies any mistake or shortcoming in the response, indicating that the constraint is not fully followed, the judgment should be: "[[The AI assistant's response does not follow the constraint]]".
\newline
\newline
Remember the following points:
\newline
(1) There should not be situations such as "partially follow the constraint". 
As long as the explanation identifies any mistakes or shortcomings in the response that result in the constraint not being fully followed, the judgment should be: "[[The AI assistant's response does not follow the constraint]]."
\newline
(2) If the explanation states that the constraints and the response are unrelated and there is no need for evaluation, the judgment should be: "[[The AI assistant's response follows the constraint]]."
\newline
\newline
Please analyze the logical consistency between the explanation and judgment following the above principles. Output your analysis first, and then output either "[[Explanation and Judgment are consistent]]" or "[[Explanation and Judgment are not consistent]]" to represent your final verdict.
\newline
\newline
Here are the constraint and the critique to be judged: 
\newline
\textbf{[The Start of Constraint]}
\newline
\textcolor{red}{\{constraint\}}
\newline
\textbf{[The End of Constraint]}
\newline
\newline
\textbf{[The Start of Critique]}
\newline
\textcolor{red}{\{critique\}}
\newline
\textbf{[The End of Critique]}
 \\
\bottomrule
\end{tabular}
}
\caption{The prompt templates for cross-model verification. }
\label{tab:cross_validation}
\end{table*}
\begin{table*} [!t]
\centering
\small
\setlength{\tabcolsep}{1.6mm}{
\begin{tabular}{p{\dimexpr \linewidth-2\tabcolsep\relax}}
\toprule
You are a fair judge, specializing in evaluating the quality of an AI assistant's responses to user instructions.
\newline
\newline
We will provide you with a user instruction, an AI assistant’s response, and a specific constraint from the user instruction. 
Please first determine whether the given constraint is related to length (e.g., "The article should be no less than 800 words" or "Each title must be 8 characters long"). 
If so, please extract the segments from the AI assistant’s response that should follow this length requirement (extract the original text from the response without any modifications).
\newline
\newline
\#\# Reference Example
\newline
\textcolor{red}{\{in\_context\_examples\}}
\newline
\newline
Here are the user instruction, the constraint, and the AI assistant’s response:
\newline
\textbf{[The Start of User Instruction]}
\newline
\textcolor{red}{\{instruction\}}
\newline
\textbf{[The End of User Instruction]}
\newline
\newline
\textbf{[The Start of Constraint]}
\newline
\textcolor{red}{\{constraint\}}
\newline
\textbf{[The End of Constraint]}
\newline
\newline
\textbf{[The Start of AI Assistant’s Response]}
\newline
\textcolor{red}{\{response\}}
\newline
\textbf{[The End of AI Assistant’s Response]}
\newline
\newline
Remember the following points:
\newline
(1) The given constraints may contain length requirements for multiple segments of the response. Therefore, numerous segments of content may need to be extracted.
\newline
(2) You only need to determine whether the given constraint is related to length, without considering other constraints of the user instruction.
\newline
(3) Output your analysis and conclusion strictly following the format below:
\newline
\newline
\textbf{Analysis}: ...(Provide a detailed analysis to determine whether the given constraint is related to length)
\newline
\textbf{Conclusion}: ...
\newline
1. (If the constraint is NOT related to length)
\newline
\newline
"""json
\newline
\{
\newline
\hspace*{2em} "\textbf{Length Constraint}" : False
\newline
\}
\newline
"""
\newline
\newline
2. (If the constraint is related to length)
\newline
\newline
"""json
\newline
\{
\newline
\hspace*{2em}   "\textbf{Length Constraint}" : True,
\newline
\hspace*{2em}    "\textbf{Extracted Segments}" : [
\newline
\hspace*{4em}        \{
\newline
\hspace*{6em}      "\textbf{Length Requirement within the Constraint}": ...(Provide the length requirement within the constraint. Be sure to extract the original text from the constraint without making any modification),
\newline
 \hspace*{6em}           "\textbf{Corresponding Segment in Response}": ...(Provide the segment of the response that should follow this requirement without making any modification. If no corresponding segment exists in the response, output "No corresponding segment exists.")
\newline
 \hspace*{4em}       \},
\newline
\hspace*{4em}        \{
\newline
\hspace*{6em}           "\textbf{Length Requirement within the Constraint}": ...,
\newline
\hspace*{6em}            "\textbf{Corresponding Segment in Response}": ...
\newline
\hspace*{4em}        \},
\newline
\hspace*{4em}        ... (Analogously, extracting all segments in the response that should follow the length requirements of the given constraint)
\newline
\hspace*{2em}    ]
\newline
\}
\newline
"""
\\
\bottomrule
\end{tabular}
}
\caption{The prompt template for length constraint identification and segment extraction. }
\label{tab:length_constraint_identification}
\end{table*}
\begin{table*} [!t]
\centering
\small
\setlength{\tabcolsep}{1.6mm}{
\begin{tabular}{p{\dimexpr \linewidth-2\tabcolsep\relax}}
\toprule
You are a fair judge, specializing in evaluating the quality of an AI assistant's responses to user instructions.
\newline
\newline
We will provide you with a user instruction, a specific constraint from the user instruction, and the AI assistant’s response. Please think carefully
and provide a detailed analysis of whether the AI assistant’s response follows the given constraint. 
\newline
\newline
The given constraint is related to length. We will also provide a JSON data, which contains multiple items. 
Each item consists of three elements: 
(1) a length requirement within the constraint, 
(2) a segment of the responses that should follow the length requirement, and (3) the actual length of the segment. 
It is important that you consider the "actual length provided in the JSON data" as the golden-truth length, which is absolutely accurate. 
Do not make your own calculation about “the length of the segment”.
You must rely solely on the "actual length" as the basis for length validation.
Output your explanation and judgment for the given constraint based on the actual length provided in the JSON data, strictly following the format below:
\newline
\newline
"""
\newline
\textbf{[The Start of Constraint]}
\newline
\textbf{Constraint:} ... (Directly output the first constraint from the checklist here with no modifications)
\newline
\textbf{Explanation:} ... (Provide a thorough, step-by-step analysis of whether the AI assistant's response follows this constraint,  referencing specific details of the response)
\newline
\textbf{Judgment:} [[The AI assistant’s response follows this constraint]] or [[The AI assistant’s response does not follow this constraint]] 
\newline
\textbf{[The End of Constraint]}
\newline
"""
\newline
\newline
Remember the following points:
\newline
(1) You should only focus on whether the response follows the constraint, not on the overall quality of the response. Other shortcomings in the response (such as not fulfilling the constraint to a high standard) should not affect your judgment unless the constraint is not followed.
\newline
(2) Your judgment should be as strict as possible. You should output "[[The AI assistant’s response follows this constraint]]" only if the response completely follows every part of the constraint as stated, without any omission or mistake. 
\newline
(3) Independently analyzing the given constraint, and does not take into account whether other constraints are followed.
\newline
(4) The given constraint may contain both length requirements and other requirements. For length requirements, you must rely solely on the "actual length provided in the JSON data" as the basis for length validation. For other requirements, please refer to the above principles and carefully evaluate them by yourself.
\newline
(5) When generating your explanation, it is forbidden to mention phrases such as "according to the given JSON data" or "based on the provided actual length".
Your explanation should convey that the length is calculated independently by you, without referencing the given JSON data.
\newline
\newline
Here are the user instruction, the constraint, the AI assistant's response, and the JSON data about length:
\newline
\textbf{[The Start of User Instruction]}
\newline
\textcolor{red}{\{instruction\}}
\newline
\textbf{[The End of User Instruction]}
\newline

\textbf{[The Start of the Constraint]}
\newline
\textcolor{red}{\{constraint\}}
\newline
\textbf{[The End of the Constraint]}
\newline

\textbf{[The Start of AI Assistant's Response]}
\newline
\textcolor{red}{\{response\}}
\newline
\textbf{[The End of AI Assistant's Response]}
\newline
\newline
\textbf{[The Start of The Json Data]}
\newline
\textcolor{red}{\{json\_data\}}
\newline
\textbf{[The End of The Json Data]}
\\
\bottomrule
\end{tabular}
}
\caption{The prompt template for rule-augmented critique revision. }
\label{tab:rule_revision}
\end{table*}
\begin{table*} [t]
\centering
\setlength{\tabcolsep}{1.6mm}{
\begin{tabular}{p{\dimexpr \linewidth-2\tabcolsep\relax}}
\toprule
Below, an instruction, a corresponding model response, and a constraint checklist for the instruction will be provided. 
The checklist contains some of the constraints within the instruction.
Your task is to verify whether the model response follows each constraint in the checklist, choose "Followed" or "Not Followed", and provide a brief justification.
\newline
\newline
\textbf{Task Details:}

1. Please analyze whether the response follows each constraint listed in the given checklist, providing a judgment for each constraint respectively.
\newline
2. Your judgments must be strict. Only responses that fully satisfy a constraint can be judged as "Followed". If there is any omission or error regarding a constraint, it must be judged as "Not Followed".
\newline
3. Please focus exclusively on the constraints within the given checklist. It is unnecessary to consider whether the response follows any other constraints beyond the checklist.
\newline
4. When judging the following of each constraint, your judgement should consider the complete context of the instructions, rather than interpreting the constraint in isolation. 
\newline

{\color{red} \{in-context examples\}}\newline

\textbf{[Instruction]}\newline
{\color{red} \{instruction\}}\newline

\textbf{[Model Response]}\newline
{\color{red} \{model\_response\}}\newline

\textbf{[Constraint Checklist]}\newline
{\color{red}\{checklist\}}\newline

Your choice for the first constraint in the checklist: {\color{blue} \{option\}} \quad A. Followed\quad B. Not Followed
\newline
Your justification:  {\color{blue} \{justification\}}
\newline
\newline
Your choice for the second constraint in the checklist: {\color{blue} \{option\}} \quad A. Followed\quad B. Not Followed
\newline
Your justification:  {\color{blue} \{justification\}}
\newline
\newline
……
\\
\bottomrule
\end{tabular}
}
\caption{Human annotation guideline for constraint verification. 
The {\color{red} red} part is the information provided to the annotators, and the {\color{blue} blue} part is content that requires the annotators to make annotations.}
\label{tab:human_annotation}
\end{table*}

\clearpage

\end{document}